%% file: main.tex
\documentclass{MITcsail}
\usepackage{amsmath}

\usepackage{microtype}
\usepackage{enumitem}
\usepackage[capitalize,noabbrev]{cleveref}
\usepackage[table]{xcolor}
\definecolor{lightgreen}{RGB}{220,245,220}
\definecolor{codegreen}{rgb}{0,0.6,0}
\definecolor{codegray}{rgb}{0.5,0.5,0.5}
\definecolor{codepurple}{rgb}{0.58,0,0.82}
\definecolor{backcolour}{rgb}{0.95,0.95,0.92}
\usepackage{fancyhdr}

\usepackage{listings}
\lstdefinestyle{mystyle}{
    backgroundcolor=\color{backcolour},
    commentstyle=\color{codegreen},
    keywordstyle=\color{blue},
    numberstyle=\tiny\color{codegray},
    stringstyle=\color{codepurple},
    basicstyle=\ttfamily\footnotesize,
    breakatwhitespace=false,
    breaklines=true,
    captionpos=b,
    keepspaces=true,
    numbers=left,
    numbersep=5pt,
    showspaces=false,
    showstringspaces=false,
    showtabs=false,
    tabsize=2
}

\usepackage{tikz}
\newcommand{\circled}[1]{%
  \tikz[baseline=(C.base)]\node[draw, circle, inner sep=0.6pt] (C) {\footnotesize #1};%
}
\newcommand{\cpt}{\fontsize{18.5pt}{20pt}\selectfont}

\usepackage[table]{xcolor} 
\usepackage[textsize=tiny]{todonotes}

\newif\ifcomments
\newcommand{\eat}[1]{}
\commentstrue
\ifcomments
    \providecommand{\shubham}[1]{{\color{red}{/* shubham: #1 */}}}
    \providecommand{\mert}[1]{{\color{olive}{/* mert: #1 */}}}
    
    \providecommand{\accheng}[1]{{\color{blue}{/* accheng: #1 */}}}

\else
    \providecommand{\shubham}[1]{}
    \providecommand{\mert}[1]{}
    \providecommand{\accheng}[1]{}
\fi

\usepackage{xspace}
\newcommand{\sys}{EvoX\xspace}

\usepackage{listings}
\usepackage{xcolor}
\usepackage{booktabs}
\usepackage{tabularx}
\usepackage{float}

\definecolor{codegreen}{rgb}{0,0.6,0}
\definecolor{codegray}{rgb}{0.5,0.5,0.5}
\definecolor{codepurple}{rgb}{0.58,0,0.82}
\definecolor{backcolour}{rgb}{0.95,0.95,0.92}
\definecolor{mintedbg}{RGB}{242,242,242}    
\definecolor{mintedframe}{RGB}{100,100,100} 

\usepackage[most]{tcolorbox}
\tcbuselibrary{skins,breakable}

\definecolor{archbg}{RGB}{245,248,252}
\definecolor{archframe}{RGB}{90,120,160}
\definecolor{archaccent}{RGB}{70,100,140}
\newtcolorbox{configblock}[1]{
  breakable,
  enhanced,
  colback=archbg,
  colframe=archframe,
  boxrule=0.9pt,
  arc=2.5mm,
  left=1.4mm,right=1.4mm,top=1.0mm,bottom=1.0mm,
  title=\textbf{#1},
  fonttitle=\small,
}

\usepackage{algorithm}
\usepackage{algpseudocode}

\usepackage[textsize=tiny]{todonotes}
\usepackage{booktabs}
\usepackage{xspace}
\usepackage{authblk}
\title{\textcolor[HTML]{002676}{\cpt EvoX: Meta-Evolution for Automated Discovery}}

\author{%
  \begin{minipage}{\textwidth}
    \raggedright
    \vspace{-0.1ex}
    {\fontsize{11pt}{12pt}\selectfont\normalfont\mdseries
    
    Shu Liu$^{1*}$, Shubham Agarwal$^{1*}$, Monishwaran Maheswaran$^{1}$, Mert Cemri$^{1}$, Zhifei Li$^{1}$, Qiuyang Mang$^{1}$, Ashwin Naren$^{1}$, Ethan Boneh$^{2}$, Audrey Cheng$^{1}$, Melissa Z. Pan$^{1}$, Alexander Du$^{1}$, Kurt Keutzer$^{1}$,
    Alvin Cheung$^{1}$,
    Alexandros G. Dimakis$^{1,3}$, Koushik Sen$^{1}$, Matei Zaharia$^{1}$, Ion Stoica$^{1}$\\
    \vspace{1ex}}
    \vspace{1ex}    

    \textbf{\small Affiliations}\\
    \vspace{1ex}
    {\small $^{1}$UC Berkeley \quad $^{2}$Stanford University \quad $^{3}$Bespoke Labs 
    \\}
  \end{minipage}%
}
\makeatletter
\providecommand{\pagehead}{EvoX: Meta-Evolution for Automated Discovery} 
\makeatother

\begin{document}
\maketitle

\input{tex/abstract}
\input{tex/intro}

\input{tex/background}

\input{tex/method}

\input{tex/new_case_study}

\input{tex/exp}

\input{tex/conclusion}

\bibliographystyle{plain}
\bibliography{references}

\beginsupplement
\appendix

\input{tex/appendix}
\end{document}

%% file: tex/abstract.tex
\begin{abstract}
\vspace{-1em}
\textbf{Abstract:} Recent work such as AlphaEvolve has shown that combining LLM-driven optimization with evolutionary search can effectively improve programs, prompts, and algorithms across domains. In this paradigm, previously evaluated solutions are reused to guide the model toward new candidate solutions. Crucially, the effectiveness of this evolution process depends on the search strategy: how prior solutions are selected and varied to generate new candidates. However, most existing methods rely on fixed search strategies with predefined knobs (e.g., explore–exploit ratios) that remain static throughout execution. 
While effective in some settings, these approaches often fail to adapt across tasks, or even within the same task as the search space changes over time.

We introduce EvoX, an adaptive evolution method that optimizes its own evolution process. EvoX jointly evolves candidate solutions and the search strategies used to generate them, continuously updating how prior solutions are selected and varied based on progress.
This enables the system to dynamically shift between different search strategies during the optimization process.
Across nearly 200 real-world optimization tasks, EvoX outperforms existing AI-driven evolutionary methods including AlphaEvolve, OpenEvolve, GEPA, and ShinkaEvolve on the majority of tasks.

\end{abstract}

%% file: tex/intro.tex
\begin{table}[H]
\centering
\footnotesize
\setlength{\tabcolsep}{2pt}

\caption{\textbf{\sys across optimization tasks.}
We compare \sys against strong human-designed methods and prior AI discovery systems, including AlphaEvolve, GEPA, ShinkaEvolve, and OpenEvolve, across mathematical optimization, systems performance optimization (e.g., GPU-to-model placement), and algorithm engineering tasks (e.g., Frontier-CS). For Frontier-CS, we report median scores over 172 competitive programming problems.
Arrows indicate whether higher ($\uparrow$) or lower ($\downarrow$) scores are better.
All \emph{open} AI frameworks are evaluated under a fixed budget of 100 iterations, and the best result among them is reported as “AI Best.”
AlphaEvolve results are taken directly from its original publication.}
\label{tab:first-table}
\vspace{-0.2em}

\begin{tabular}{p{3cm}ccc}
\toprule
\textbf{Task} 
& \textbf{Human Best} 
& \textbf{AlphaEvolve} 
& \textbf{\sys} \\
\midrule

\multicolumn{4}{l}{\textsc{\textbf{Mathematics}}} \\

Circle-Packing ($\uparrow$)
& 2.634
& 2.635
& \textbf{2.636} \\

MinMaxMinDist ($\downarrow$)
& 4.16584
& 4.16579
& \textbf{4.16578} \\


\addlinespace[0.2em]
\midrule
\textbf{Task}
& \textbf{Human Best}
& \textbf{AI Best}
& \textbf{EvoX} \\
\midrule

\multicolumn{4}{l}{\textsc{\textbf{Systems Optimization}}} \\

Cloud Transfer ($\downarrow$)
& 626.24
& 645.72
& \textbf{623.69} \\

GPU Placement ($\uparrow$)
& 21.89
& 26.26
& \textbf{30.52} \\

Txn Scheduling ($\uparrow$)
& 2724.8
& 4329
& \textbf{4347.8} \\

\addlinespace[0.4em]
\multicolumn{4}{l}{\textsc{\textbf{Algorithms}}} \\


Frontier-CS ($\uparrow$)
& --
& 56.2
& \textbf{75.5} \\


\bottomrule
\end{tabular}
\end{table}
\noindent\rule{\linewidth}{0.4pt}
{\footnotesize
\noindent * denotes equal contribution. Code is available at \url{https://github.com/skydiscover-ai/skydiscover}.
}

\vspace{-2em}
\newcounter{subfig}
\renewcommand{\thesubfig}{\alph{subfig}}

\begin{figure*}[t]
    \centering

    \refstepcounter{subfig}
    \begin{minipage}[b]{0.7\textwidth}
        \centering
        \includegraphics[width=\textwidth]{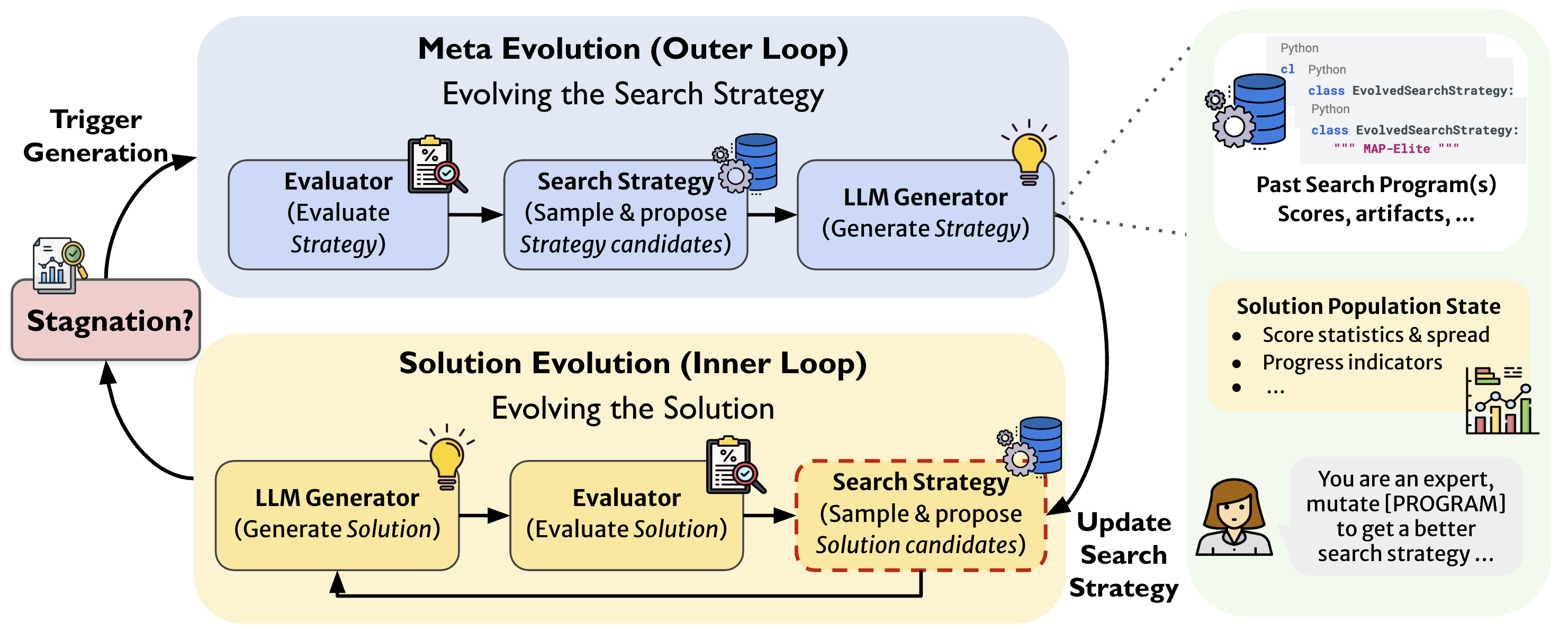}
        \vspace{-0.6em}
        \centerline{\footnotesize(\alph{subfig}) Overall system architecture}
        \label{fig:arch_a}
    \end{minipage}
    \hspace{0em}
    \refstepcounter{subfig}
    \begin{minipage}[b]{0.27\textwidth}
        \centering
        \includegraphics[width=\textwidth]{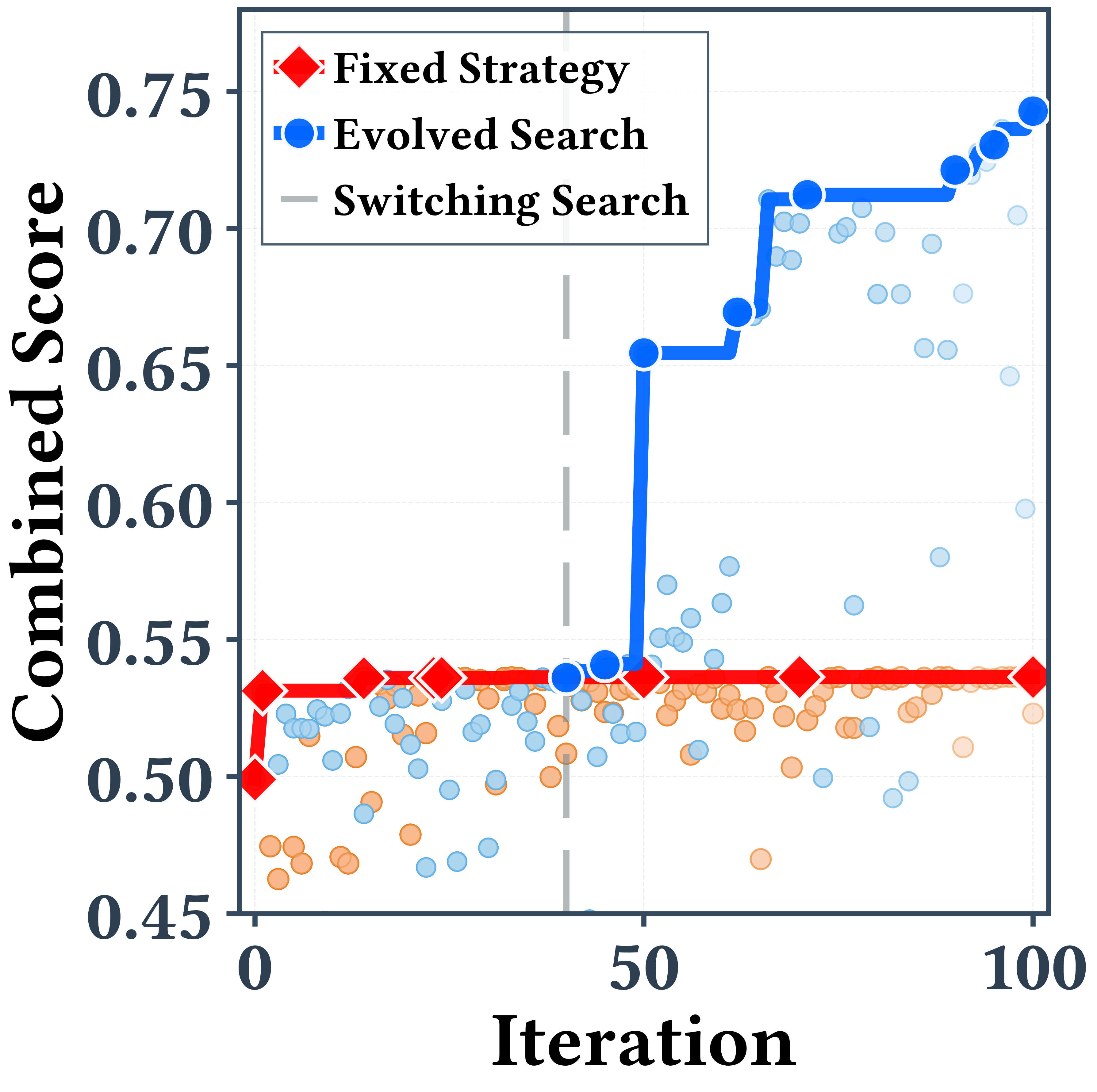}
        \vspace{-0.6em}
        \centerline{\footnotesize(\alph{subfig}) Evolved search (blue) breakthroughs %
        \vphantom{Overall system architecture}}
        \label{fig:arch_b}
    \end{minipage}
    \vspace{-1em}
    \caption{
\textbf{Evolving the search strategy.}
(\ref{fig:arch_a}) \emph{System architecture.}
\sys has two coupled loops: an inner loop that evolves solutions,
and an outer loop that evolves the \emph{search strategy} that governs generation.
(\ref{fig:arch_b}) \emph{Effect of search evolution.}
A strategy with fixed exploration-exploitation ratio (MAP-Elites, red) stagnates,
while an evolving search strategy (blue) produces discrete performance breakthroughs.
} 
    \label{fig:architecture}
    \vspace{-1em}
\end{figure*}

\section{Introduction}
LLM-driven optimization combined with evolutionary search has enabled numerous scientific breakthroughs across domains including mathematics~\cite{alphaevolve}, systems performance optimization~\cite{cheng2025barbarians}, and competitive programming~\cite{imajuku2025ale}.
Typically, LLM-driven evolutionary systems maintain a population of candidate solutions. At each step, a \textit{search strategy} selects a subset of previously evaluated solutions and constructs a prompt, which is passed to a generator model (typically an LLM) to produce new candidates. These candidates are then evaluated and added back to the population.

Critically, the effectiveness of the evolution process relies on the search strategy, which determines both (i) which candidates are selected from the population and (ii) how new solutions are proposed from them (e.g., via refinement, structural variation, or combining multiple ideas). 
These choices directly influence what the generator attempts next and which regions of the solution space to explore.

Existing LLM-driven evolutionary systems often rely on fixed search strategies with hand-specified parameters.
For example, AlphaEvolve~\cite{alphaevolve} employs MAP-Elites with predefined population database structures and selection ratios. OpenEvolve~\cite{openevolve} similarly relies on static elite and diversity heuristics.
ShinkaEvolve~\cite{shinka} takes a step toward adaptivity by incorporating bandit-based selection on LLM generators.
However, key search strategy knobs remain manually configured. For example, the exploitation ratio is fixed, controlling how often top solutions are refined.



In practice, a fixed search strategy often fails to generalize across problems or across different stages of optimization. This can lead to stagnation during the search and requires manual retuning to make progress.
Some tasks benefit primarily from repeated refinement of a strong candidate, where small local modifications incrementally improve an existing solution.
For example, in packing problems~\cite{alphaevolve, georgiev2025mathematical}, adjusting a few placements in a near-feasible configuration can yield further gains.
In contrast, other tasks require qualitatively different solution structures.
For instance, in a GPU-model placement problem, refining a least-loaded heuristic quickly plateaus, and further progress requires reformulating the problem as a bin-packing assignment.

Even within a single optimization run, the effectiveness of a search strategy can change over time. In a multi-objective signal processing task (Figure~\ref{fig:architecture}(\ref{fig:arch_b})), a MAP-Elites style strategy (red line) makes rapid early progress but later stagnates. Switching to a strategy that explicitly samples candidates along different trade-off objectives (blue line) enables continued improvement.


Motivated by this observation, \sys frames LLM-driven optimization as a meta-learning problem in which \emph{the search strategy itself is treated as an evolvable object}. Rather than fixing a hand-tuned search strategy, \sys operates as a two-level evolution process comprising a \emph{solution-evolution} loop and a \emph{meta-evolution} loop (Figure~\ref{fig:architecture}(\ref{fig:arch_a})).

\textbf{The solution-evolution loop} generates candidate solutions under the standard LLM-driven evolutionary search paradigm.
At each step, a search strategy selects prior candidates and determines how to generate new ones, for example by sampling high-performing solutions and applying a variation operator (e.g., refinement or structural variation).
The resulting prompt is passed to the LLM, which produces a new candidate that is evaluated and added back to the population.


\textbf{The meta-evolution loop} periodically updates the search strategy. Each strategy is deployed for a window of solution-evolution iterations and evaluated by the progress it induces on the downstream task (e.g., improvement rate). 
When progress stagnates, the meta-evolution loop generates a new search strategy using the LLM, conditioned on prior strategies, their observed performance, and the current state of the solution population.
This loop enables \sys to evolve the search strategy itself: selecting, mutating, and replacing strategies based on their effectiveness.
Because the performance of a strategy depends on the evolving population, \sys explicitly conditions strategy generation on population-level signals, allowing it to adapt as the search space changes.

\paragraph{Contributions.} We evaluate \sys across nearly 200 real-world optimization tasks spanning mathematics, algorithms, and scientific research benchmarks (Table~\ref{tab:first-table} and Section~\ref{sec:eval}).
Starting from a simple random-sampling search strategy, \sys consistently outperforms existing LLM-driven evolutionary frameworks, including OpenEvolve, ShinkaEvolve, and GEPA~\cite{openevolve, shinka, gepa} on the majority of tasks (e.g., 96\% of the math and system optimization benchmarks). It also often matches or surpasses the best human-designed solutions. In summary, we make the following contributions:

\begin{enumerate}
\item We formalize LLM-driven optimization as a two-level process that separates solution evolution from search strategy evolution.

\item We introduce \sys, a method that dynamically evolves search strategies based on improvements in the optimization objective

\item We demonstrate consistent improvements over prior methods across nearly 200 problems, and characterize the cost, scaling behavior, and adaptation dynamics of the search strategy evolution.
\end{enumerate}

%% file: tex/background.tex
\section{Related Work}

\paragraph{Context and Memory Management.}

Many recent LLM systems iteratively improve solutions by incorporating feedback from prior attempts. 
A key challenge in this setting is how to store, summarize, and present past information to guide future generations.
ACE~\cite{zhang2025agentic} treats context as editable objects, MemGPT~\cite{packer2023memgpt} introduces hierarchical memory, and Reflexion~\cite{reflexion} and Self-Refine~\cite{madaan2023self} incorporate model-generated feedback into future prompts.
These approaches improve how prior experience is organized and reused, enabling more effective iterative refinement.

One prominent paradigm that builds on this foundation is evolutionary search, which treats prior solutions as a population and explicitly selects and varies candidates over time.

\paragraph{LLM-Guided Evolutionary Search.}
Recent work applies evolutionary search to guide LLM-driven optimization, spanning prompt optimization~\cite{fernando2023promptbreeder, guo2023evoprompt, lee2025evolving, ye2024reevo, suzgun2025dynamic, chopra2025feedback} and program or algorithm discovery~\cite{alphaevolve, openevolve, shinka, gepa, assumpccao2025codeevolve, hemberg2024evolving, shojaee2024llm}. 
These systems differ primarily in how candidates are selected and varied.
AlphaEvolve~\cite{alphaevolve} selects candidates using MAP-Elites, GEPA~\cite{gepa} selects along Pareto frontiers, and OpenEvolve~\cite{openevolve} and ShinkaEvolve~\cite{shinka} emphasize diversity-driven selection. 
CodeEvolve~\cite{assumpccao2025codeevolve} integrates LLM-based generation within an island-based genetic algorithm, while DeltaEvolve~\cite{jiang2026deltaevolve} models semantic deltas between candidates. 
PACEvolve~\cite{yan2026pacevolve} introduces mechanisms such as hierarchical context pruning and momentum-based backtracking to improve stability. However, these systems still operate under predefined search strategies with manually designed knobs.

Some recent work introduces learning-based adaptation mechanisms. SOAR~\cite{pourcel2025self} alternates between search and model fine-tuning, while ThetaEvolve~\cite{theta}, TTT-Discover~\cite{yuksekgonul2026learning}, and FLEX~\cite{cai2025flex} apply reinforcement learning to improve the generator model. 
These approaches adapt the generator model itself, but do not adapt the search strategy governing candidate selection and variation.

\paragraph{Meta-Learning and Learning to Optimize.}
Meta-learning and learned optimization frameworks treat the optimization procedure itself as the object of adaptation rather than a fixed component~\cite{metz2019understanding, chen2022learning}. Prior work explores learned optimizers, symbolic discovery of update rules~\cite{chen2023symbolic}, gradient-based meta-learning~\cite{andrychowicz2016learning}, and reinforcement learning-based optimizers~\cite{metarloptimizer}. 
These works demonstrate that adapting the optimization process itself can significantly improve search efficiency.
Our work extends this principle to LLM-driven evolutionary search.

\paragraph{EvoX: Meta-Evolving the Search Strategy.}
\sys builds on these lines of work by treating the search strategy itself as an evolvable object. 
Rather than relying on fixed candidate selection and variation mechanisms, EvoX evolves the search strategy through evolutionary feedback, dynamically adapting how candidates are selected and varied across different optimization stages and heterogeneous solution landscapes.

%% file: tex/method.tex
\section{Problem Formulation}
\label{sec:problem}

We formalize LLM-driven evolutionary search following standard abstractions from prior work~\cite{FunSearch2024,alphaevolve,openevolve,gepa,shinka}. Candidate solutions are generated by a language model, evaluated by a task-specific evaluator, and iteratively evolved under a search strategy and a fixed evaluation budget.

\paragraph{Candidate solution evaluation.} Let $\mathcal{X}$ denote the space of candidate solutions (e.g., programs or prompts).
Each candidate $x \in \mathcal{X}$ is evaluated by an evaluator
\[
E(x) \rightarrow \bigl(s(x),\, a(x)\bigr),
\]
which returns a scalar score $s(x)\in\mathbb{R}$ together with auxiliary artifacts $a(x)$
(e.g., logs, traces, or any feedback).

\paragraph{Solution population database.}
The optimization proceeds for $T$ sequential evaluation steps.
Let
\[
\mathcal{D}_t = \{(x_i, s_i, a_i)\}_{i=1}^{t},
\quad \mathcal{D}_0=\emptyset,
\]
denote the database of all candidate solutions evaluated up to step $t$.
At each step, a new candidate is generated, evaluated, and appended to the database to form $\mathcal{D}_{t+1}$.

\paragraph{Search strategy.}
A search strategy $S \in \mathcal{S}$ specifies how the next-generation LLM input
is constructed from the current database $\mathcal{D}_t$.
Concretely, $S$ defines the construction of the generation context:
\[
C_S(\mathcal{D}_t) \rightarrow \bigl(x_{\mathrm{par}},\, \pi,\, \mathcal{I}\bigr),
\]
which selects (i) one or more \emph{parent} candidate(s) $x_{\mathrm{par}} \in \mathcal{D}_t$ to be modified,
(ii) a \emph{variation operator} $\pi$ expressed in the prompt that specifies how the
parent can be modified, and
(iii) optionally, an \emph{inspiration set} $\mathcal{I}\subseteq \mathcal{D}_t$
(e.g., diverse candidates) that provides additional exemplars. 
This abstraction aligns with prior LLM-driven evolutionary systems~\cite{shinka, openevolve}.


Given $(x_{\mathrm{par}}, \pi, \mathcal{I})$, the solution generator (implemented by an LLM) produces a new candidate
\[
x' \sim \mathcal{G}_{\mathrm{sol}}(\cdot \mid x_{\mathrm{par}}, \pi, \mathcal{I}),
\]
which is evaluated and appended to the database.

\emph{Variation operators.}
The operator $\pi$ specifies the \emph{kind} of modification requested in a generation step.
We use three variation operators: \circled{1} \textbf{local refinement} for fine-grained edits (exploitation),
\circled{2} \textbf{structural variation} for coarse-grained redesigns (exploration), and
\circled{3} \textbf{free-form variation}, which imposes no constraints on the edit scope. 


The semantics of these operators are task-dependent.
For example, in code or algorithmic tasks, refinement may tune parameters, reorder logic,
or edit localized code blocks, whereas structural variation may switch algorithm families or restructure the overall design.
At the start of each problem, we instantiate these operators using a lightweight model
(Section~\ref{sec:eval}) to generate a small set of operator-specific prompts from the
problem description.
During optimization, the search strategy selects among these operators to
control the intended type of variation applied to the parent candidate.

\paragraph{Optimization goal.}
Search strategies shape optimization behavior, and our goal is to
\textit{adaptively select and improve the search strategy} to maximize
the final best score
\[
\max_{(x,s,a)\in \mathcal{D}_T} s
\]
under a fixed evaluation budget of $T$ steps.

\section{Co-evolving Solution and Search Strategy}
\label{sec:method}

\input{table/algorithm}

\sys addresses the problem of iteratively improving solution 
quality under a fixed evaluation budget defined in Section~\ref{sec:problem} by \emph{co-evolving} two components:
(i) the solution population $\mathcal{D}_t$, and
(ii) the search strategy $S$ that constructs the next-generation LLM context.
Specifically, \sys proceeds through the three-step process shown in Algorithm~\ref{alg:evosquared}: (1) evolving the solution database under a fixed search strategy, (2) monitoring population performance, and (3) updating the search strategy when progress stalls, using feedback from
previous strategy deployments and the current state of the solution population.

\subsection{Solution evolution under the current strategy}
\label{sec:method-solution}
Given the current database $\mathcal{D}_t$ and an active strategy $S_t$,
\sys generates and evaluates new candidates as defined in
Section~\ref{sec:problem}:
the strategy constructs a generation context
$(x_{\mathrm{par}}, \pi, \mathcal{I}) \sim C_{S_t}(\mathcal{D}_t)$,
the generator produces $x'$, and the evaluator appends $(x', s', a')$ to the database.

A search strategy controls both \emph{what} the model sees (through parent selection and inspiration construction) and \emph{how} the selected parent is transformed (via the variation operator $\pi$).

\subsection{Progress monitoring and strategy updates}
\label{sec:method-progress}
A search strategy does not affect a single candidate in isolation, but shapes
a \emph{sequence} of generated candidates.
Accordingly, its effectiveness can only be assessed over multiple evaluation
steps rather than from a single outcome.
For this reason, \sys monitors progress over a sliding window of the most recent
$W$ evaluation steps.

Let $t$ denote the start of the current monitoring window and define
\begin{equation}
s_{\mathrm{start}} = \max_{(x,s,a)\in\mathcal{D}_t} s, \quad
s_{\mathrm{end}} = \max_{(x,s,a)\in\mathcal{D}_{t+W}} s, \quad
\Delta = s_{\mathrm{end}} - s_{\mathrm{start}} .
\end{equation}
We treat $\Delta$ as the primary signal of strategy efficacy.
If $\Delta$ falls below a stagnation threshold $\tau$, \sys triggers a strategy update; otherwise, it continues with the current strategy.
This design makes strategy switching \emph{demand-driven} rather than \emph{periodic at a fixed interval}, avoiding unnecessary updates when the current strategy is effective.

\paragraph{Search strategy evaluation.} After $W$ evaluation steps have elapsed, \sys computes a performance score for the strategy employed during that window: 
\begin{equation}
\begin{aligned}
J(S_t \mid \mathcal{D}_t)
&=
\frac{
(s_{\mathrm{end}} - s_{\mathrm{start}})
\,\log(1 + s_{\mathrm{start}})
}{
\sqrt{W}
}
\end{aligned}
\label{eq:strategy_score}
\end{equation}

The $\log(1+s_{\mathrm{start}})$ term upweights improvements achieved from higher starting scores, rewarding strategies that drive progress near the frontier where gains are typically harder to obtain. The $\sqrt{W}$ normalization accounts for window length.


\subsection{Meta-evolving the search strategy}
\label{sec:method-meta}

Adapting search strategies is challenging because their effectiveness is \emph{state-dependent}. For instance, a strategy that drives rapid progress at one stage of search may become ineffective as the solution population evolves. This arises from the inherently non-stationary nature of evolutionary optimization: as the database grows, the distribution of candidate quality, diversity, and variation outcomes shifts, altering which search behaviors are beneficial.

\sys addresses this challenge by conditioning strategy updates on both \emph{a population of past search strategies} and \emph{the current downstream solution population state}. Conditioning on the population descriptor $\phi(D_t)$ allows the system to interpret stagnation relative to population structure (e.g., loss of diversity, repeated parent selection, ineffective variation). Conditioning on the strategy database $H$ provides empirical evidence about which strategies previously induced progress under similar states.

Together, these signals enable state-conditional strategy adaptation rather than fixed or periodic strategy switching.

\paragraph{Search strategy database.} To achieve this, \sys maintains a \emph{search strategy database}
\[
\mathcal{H} = \{(S_j, \phi_j, J_j)\}_{j=1}^{M},
\]
which serves as a memory of previously deployed strategies.
Each entry records a strategy $S_j$, a descriptor
$\phi_j = \phi(\mathcal{D}_{t_j})$ summarizing the population state before and after deployment,
and its observed performance $J_j$ computed using Eq.~\ref{eq:strategy_score}.
This database enables \sys to reason about \emph{which strategies tend to work under
which search conditions}.

\emph{Population state descriptor.}
The descriptor $\phi(\mathcal{D}_t)$ summarizes the current state of the solution
population.
It includes: (i) score statistics (best value, percentiles, spread),
(ii) frontier structure (e.g., top-$k$ scores),
(iii) progress indicators (e.g., steps since last significant improvement),
and (iv) recent window statistics (e.g., parent selection frequency).

\paragraph{Meta-evolution of search strategies.}
When a strategy update is triggered, \sys evolves a new search strategy by applying variation to high-performing strategies from the history of the search.
The strategy database $\mathcal{H}$ plays the role of a population, where each
individual strategy $S_j$ is associated with a score signal $J_j$ evaluated
in the population state $\phi_j$ in which it was deployed.

\sys performs score-biased selection over
$\mathcal{H}$ to choose a high-performing parent strategy to mutate, and selects an inspirational set of strategies prioritizing those that previously induced strong progress and those that performed well under similar population descriptors.
The strategy generator (i.e., an LLM), denoted $\mathcal{G}_{\mathrm{str}}$, then applies mutation to the
selected parent, conditioned on the current population descriptor
$\phi(\mathcal{D}_t)$, to produce a new strategy candidate $S'$:
\[
S' \sim \mathcal{G}_{\mathrm{str}}\!\left(\,\cdot \mid S_{\mathrm{par}},
\phi(\mathcal{D}_t)\right),
\]

Mutations modify the components of a strategy (e.g., parent selection
rules, construction of the inspiration set, or preferences over variation
operators $\pi$).

\paragraph{Strategy deployment.}
Because strategies directly affect execution, \sys validates each candidate strategy
before deployment.
If validation succeeds, \sys immediately switches to $S'$; otherwise, it retries
generation up to a fixed budget and falls back to the previous strategy if all
attempts fail.
Importantly, switching strategies never resets the solution population:
the database $\mathcal{D}$ is preserved, and evolution continues from the current
search state.

%% file: table/algorithm.tex
\begin{algorithm}[t]
\caption{\sys: Two-level evolution process}
\label{alg:evosquared}
\small
\begin{algorithmic}[1]
\Require Budget $T$, window $W$, stagnation threshold $\tau$
\Require Evaluator $E$; solution generator $\mathcal{G}_{\mathrm{sol}}$; strategy generator $\mathcal{G}_{\mathrm{str}}$
\Require Population descriptor $\phi(\cdot)$; validity test $\textsc{Valid}(\cdot)$
\Require Initial solution database $\mathcal{D}_0$; initial strategy $S_0$
\State $\mathcal{D}_t \gets \mathcal{D}_0$;\; $S_t \gets S_0$;\; $\mathcal{H} \gets \emptyset$;\; $t \gets 0$
\While{$t < T$}
    \Statex \vspace{0.3em}\textbf{Phase I: Solution evolution under $S_t$ (one window)}
    \State $\phi_t \gets \phi(\mathcal{D}_t)$
    \State $s_{\mathrm{start}} \gets \max_{(x,s,a)\in \mathcal{D}_t} s$
    \For{$i=1$ to $W$}
        \If{$t \ge T$} \textbf{break} \EndIf
        \State $(x_{\mathrm{par}}, \pi, \mathcal{I}) \sim C_{S_t}(\mathcal{D}_t)$
        \State $x' \sim \mathcal{G}_{\mathrm{sol}}(\cdot \mid x_{\mathrm{par}}, \pi, \mathcal{I})$
        \State $(s', a') \gets E(x')$
        \State $\mathcal{D}_{t+1} \gets \mathcal{D}_t \cup \{(x', s', a')\}$;\; $t \gets t+1$;\; $\mathcal{D}_t \gets \mathcal{D}_{t+1}$
    \EndFor
    \Statex \vspace{0.3em}\textbf{Phase II: Progress monitoring}
    \State $s_{\mathrm{end}} \gets \max_{(x,s,a)\in \mathcal{D}_t} s$
    \State $\Delta \gets s_{\mathrm{end}} - s_{\mathrm{start}}$
    \State $J_t \gets \Delta \log(1+s_{\mathrm{start}})/\sqrt{W}$
    \State $\mathcal{H} \gets \mathcal{H} \cup \{(S_t, \phi_t, J_t)\}$
    \Statex \vspace{0.3em}\textbf{Phase III: Strategy evolution (on stagnation)}
    \If{$\Delta < \tau$}
        \State $S' \sim \mathcal{G}_{\mathrm{str}}(\cdot \mid \mathcal{H}, \phi(\mathcal{D}_t))$
        \If{$\textsc{Valid}(S')$} $S_t \gets S'$ \EndIf
    \EndIf
\EndWhile
\State \Return $\arg\max_{(x,s,a)\in \mathcal{D}_T} s$
\end{algorithmic}
\end{algorithm}

%% file: tex/new_case_study.tex
\section{Case Study: Signal Processing}
\label{sec:case-study-signal-main}
\begin{figure}[t]
    \centering
    \includegraphics[width=0.75\textwidth]{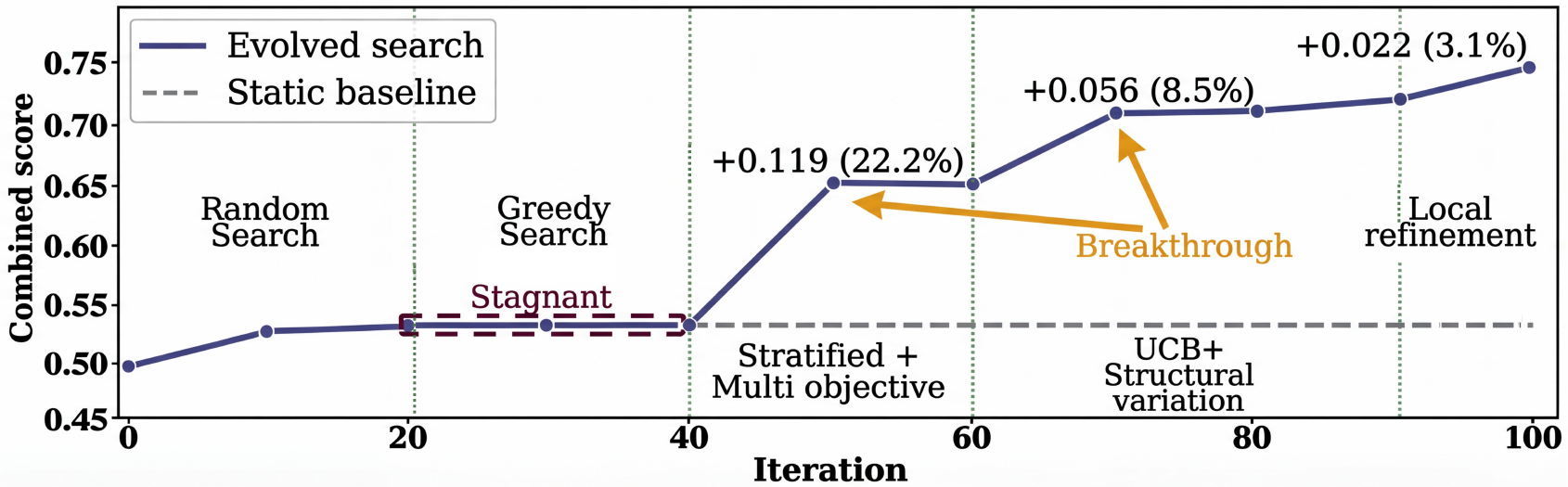}
    \vspace{-0.6em}
    \caption{
    \textbf{Evolving search strategy on the signal processing task.}
    Starting from a uniform random sampling strategy, \sys detects stagnation and adaptively
    switches to greedy search, stratified multi-objective sampling,
    UCB-guided structural variation, and finally local refinement.
    These strategy changes enable discovery of improved filtering programs
    and yield major gains at iterations $\sim$48 (+0.119), $\sim$70 (+0.056),
    and $\sim$96 (+0.022), achieving 34.1\% higher final score than the static baseline.
    }

    \label{fig:sp_case_study}
\end{figure}
We illustrate \sys through a case study on a signal processing task~\cite{shenoi2005introduction, openevolve}. The goal is to build a filtering program for a noisy, changing time series. To succeed, the program must balance several competing goals: high fidelity to the signal, smoothness to reduce noise, low lag for responsiveness, and a minimal false trend changes. We evaluate candidate programs using a combined score that balances these four objectives.

Figure~\ref{fig:sp_case_study} compares \sys against a static baseline under the
same 100-iteration budget. 
The static baseline uniformly samples the parent and inspiration set from the population and applies free-form variation to generate new candidates throughout the search.
\sys achieves a 34.1\% higher final score by
adaptively changing its search strategy based on observed progress,
as described in Section~\ref{sec:method}.

\textbf{Static Baseline.}
The dashed curve in Figure~\ref{fig:sp_case_study} illustrates the limitations of a fixed search strategy.
This strategy 
achieves modest early improvement (0.499 $\rightarrow$ 0.530) but then stagnates. 
Because the baseline chooses its parent and inspiration programs randomly and uses generic free-form variations, it mostly produces simple filters, such as basic moving averages (MA) or exponential moving averages (EMA) using NumPy. 
Later generations only make tiny adjustments to these simple settings, which are not enough to handle complex noise patterns. 
As a result, the search hits a ceiling and stays stuck.

\textbf{Phase 1: Random Search and Greedy Search.} \sys begins with the same random strategy as the baseline. 
At iteration 20, the system detects that progress has stopped and tries a greedy strategy that focuses entirely on refining the single best program found so far.
However, because the current best program still relies on simple MA/EMA structures, these tiny refinements fail to produce a breakthrough. The search remains stuck because the underlying program structure is too limited.

\textbf{Phase 2: The Breakthrough (Stratified + Multi-Objective).}
At iteration 40, $\sys$ learns from its past failure and switches to a more advanced strategy, where instead of only selecting the best overall program, it selects parents and inspiration programs from diverse score tiers and objective-specific rankings.

For example, if a parent filter reaches a high smoothness score but has a high lag at the same time, \sys selects an inspiration program that excels with lower lag. Conversely, if a parent preserves signal fidelity but leaves too much noise, it picks an inspiration program specialized in smoothing of signal. By merging these different strengths, \sys discovers novel hybrid designs such as combining singular spectrum analysis (SSA) with Whittaker smoothing.
This smart blending of complementary ideas creates the largest jump in performance (+0.119).

\textbf{Phase 3: Structural Exploration (UCB + Structural Variation).}
By iteration 60, as the progress slows again, the population state descriptor shows that many recently generated candidates have similar scores. This suggests that simple refinements or combination of ideas is no longer working.
To break this pattern, \sys evolves a policy that increases the use of structural variation to attempt bolder, more complex changes.

At the same time, it uses a UCB selection rule that encourages exploration of programs that have been largely ignored (rarely selected as parents).
The strategy continues to use multi-objective sampling for selecting the inspiration programs, as it proved effective in the previous search stage.
It also employs the \emph{structural variation operator} to encourage large, exploratory changes.
As a result, the newly generated solutions begin to use advanced SciPy tools to construct filtering pipelines, incorporating higher-order filters, smoothing 
kernels, and forward–backward filtering operations (e.g., \texttt{filtfilt}). 
This exploration of new solution families yields a significant further improvement (+0.056).

\textbf{Phase 4: Final Polishing (UCB + Local Refinement).}
By iteration 90, the search enters its final stage. Large structural changes now tend to destabilize performance rather than improve it.
Consequently, \sys shifts its strategy toward local refinement.
This involves making small, precise adjustments to the top discovered solutions while keeping the UCB rule to prevent the search from narrowing too quickly.
These fine-tuning steps provide the final gains (+0.022) and lock in the high score.

This example demonstrates the core power of $\sys$: it changes its search strategy based on what happened in previous steps and the current variety of programs it has found. By choosing the right strategy for the right moment, \sys avoids the plateaus that stall traditional methods and achieves substantially higher final performance.

%% file: tex/exp.tex
\section{Evaluation}
\label{sec:eval}
\textbf{Benchmarks.} We evaluate \sys on 196 real-world optimization tasks spanning mathematics (8), systems (6), and algorithmic and research problems (10 from ALE-Bench-Lite~\cite{imajuku2025ale} and 172 from Frontier-CS~\cite{mang2025frontiercs}) (Sections~\ref{sec:eval-math-opt}–\ref{sec:eval-ale}). 
We also report results on the ARC-AGI-2 benchmark, a widely used benchmark for evaluating reasoning and generalization capabilities (Appendix~\ref{appendix:additional-benchmarks}).
Detailed descriptions on the benchmarks are in Appendix~\ref{appendix:benchmark}. Additionally, we conduct ablation studies to analyze the cost and scaling behavior of \sys in Section~\ref{sec:ablations}.

\textbf{Baselines and Setups.} \label{sec:setup}
We compare \sys against strong LLM-driven evolutionary frameworks, including OpenEvolve~\cite{openevolve}, ShinkaEvolve~\cite{shinka}, and GEPA~\cite{gepa}.
For mathematical tasks, we additionally report human-best results and prior state-of-the-art AlphaEvolve~\cite{alphaevolve} numbers. For ADRS systems benchmarks, we report human-best results. We include further comparisons to CodeEvolve~\cite{assumpccao2025codeevolve} and ThetaEvolve~\cite{theta} in Appendix~\ref{appendix:full-results}.

For \sys, we use GPT-5 for search strategy generation with a window size of
10\% of the total iteration budget to detect stagnation and trigger strategy evolution
(Section~\ref{sec:method}).
All \sys runs start from a simple \textit{random search strategy} that samples both the parent and inspirational candidates uniformly at random.
The effect of different initial search strategies (e.g., Best-of-N, MAP-Elites) is shown in Section~\ref{sec:ablations}.
We report full results of this random search strategy in Appendix~\ref{appendix:full-random-otherbaseline-results}.

All \emph{open} frameworks, including OpenEvolve, ShinkaEvolve, GEPA, and \sys, are evaluated under a fixed budget of 100 iterations per task unless specified otherwise. 
For math tasks, human-best and AlphaEvolve~\cite{alphaevolve} results are taken from the original paper, as AlphaEvolve is not open-sourced and its iteration budget is not publicly available.

We report mean and best performance over three independent runs using two backbone models, GPT-5~\cite{openai2025gpt5} and Gemini-3.0-Pro~\cite{google2025gemini3}.
Mean scores reflect robustness across runs, while best scores capture peak solution quality.
Higher ($\uparrow$) or lower ($\downarrow$) scores indicate better performance depending on the task.

\subsection{Main Results: Math Optimization Problems}
\label{sec:eval-math-opt}

We evaluate eight mathematical optimization tasks (continuous and discrete) spanning geometric packing, extremal geometry, distance maximization, and sequence design (Table~\ref{tab:main_results}). Across these tasks, \sys achieves the strongest overall performance among open evolutionary frameworks, including OpenEvolve, ShinkaEvolve, and GEPA.

\input{table/math}

Under GPT-5, \sys attains the \emph{best} or tied-best result on 7 of 8 tasks, and under Gemini-3.0-Pro, it achieves the best result on all 8 tasks. In terms of \emph{mean} performance, \sys achieves the best score on 6 of 8 tasks under GPT-5 and 7 of 8 tasks under Gemini-3.0-Pro, demonstrating robust performance across runs. In the two cases where mean performance lags best performance (Heilbronn triangle and MinMaxMinDist $d=3$), early random initializations occasionally converge to strong local optima, limiting further improvement within the fixed 100 iteration budget we set.

Compared to the strongest previously reported results, including the closed-source AlphaEvolve, \sys matches or exceeds AlphaEvolve on 5 out of 7 tasks, such as Circle Packing, Circle Packing Rect, and MinMaxMinDist ($d=3$), within just 100 iterations. Since AlphaEvolve’s iteration budget is not publicly specified, direct cost comparison is not possible. 


Beyond aggregate scores, \sys discovers qualitatively distinct solutions across domains.
In circle packing, \sys achieves state-of-the-art performance by constructing a hexagonal-lattice core packing and refining it via constrained SLSQP optimization.
In MinMaxMinDist ($d=3$), \sys discovers a 14-point configuration in $\mathbb{R}^3$ by solving a constrained extremal-distance problem initialized from structured polyhedral seeds.
For the third autocorrelation inequality problem, \sys constructs a 1024-bin discretized function over the interval $[-0.25,\,0.25]$, computes its autoconvolution via FFT, and applies gradient-based optimization to minimize the peak magnitude.
These solutions outperform those discovered by existing evolutionary systems with fixed, manually designed search strategies.


\subsection{Main Results: System Performance Problems}
\label{sec:eval-adrs}

\input{table/adrs}

We also evaluate \sys on six real-world system optimization tasks spanning
expert placement load balancing (EPLB), GPU sharing (PRISM), LLM-driven analytics (LLM-SQL),
multi-cloud broadcast optimization (Cloudcast), transaction scheduling, and telemetry repair
(Table~\ref{tab:adrs_main_results}). \sys exceeds human-best results on all six benchmarks under Gemini-3.0-Pro.

For mean performance, \sys achieves the best score on all tasks under GPT-5 and 5 of 6 tasks under Gemini-3.0-Pro. Telemetry repair is the only case where OpenEvolve achieves a slightly higher mean. For best performance, \sys outperforms or ties all baselines on all tasks under GPT-5 and 5 of 6 tasks under Gemini-3.0-Pro.

Qualitatively, \sys uncovers new system optimization algorithms.
For example, in Cloudcast, \sys discovers a Steiner-tree–based multicast routing strategy that builds a shortest-path distance graph over regions and jointly optimizes all destinations, reducing transfer cost beyond prior heuristics.
In PRISM, \sys identifies a model-placement strategy that minimizes peak KV-cache pressure by binary-searching a global load threshold and applying Best-Fit-Decreasing packing.

\subsection{Main Results: Algorithmic and Research Problems}
\label{sec:eval-ale}

\begin{figure}[t]
    \centering
    \begin{subfigure}[t]{.48\columnwidth}
        \centering
        \includegraphics[width=\linewidth]{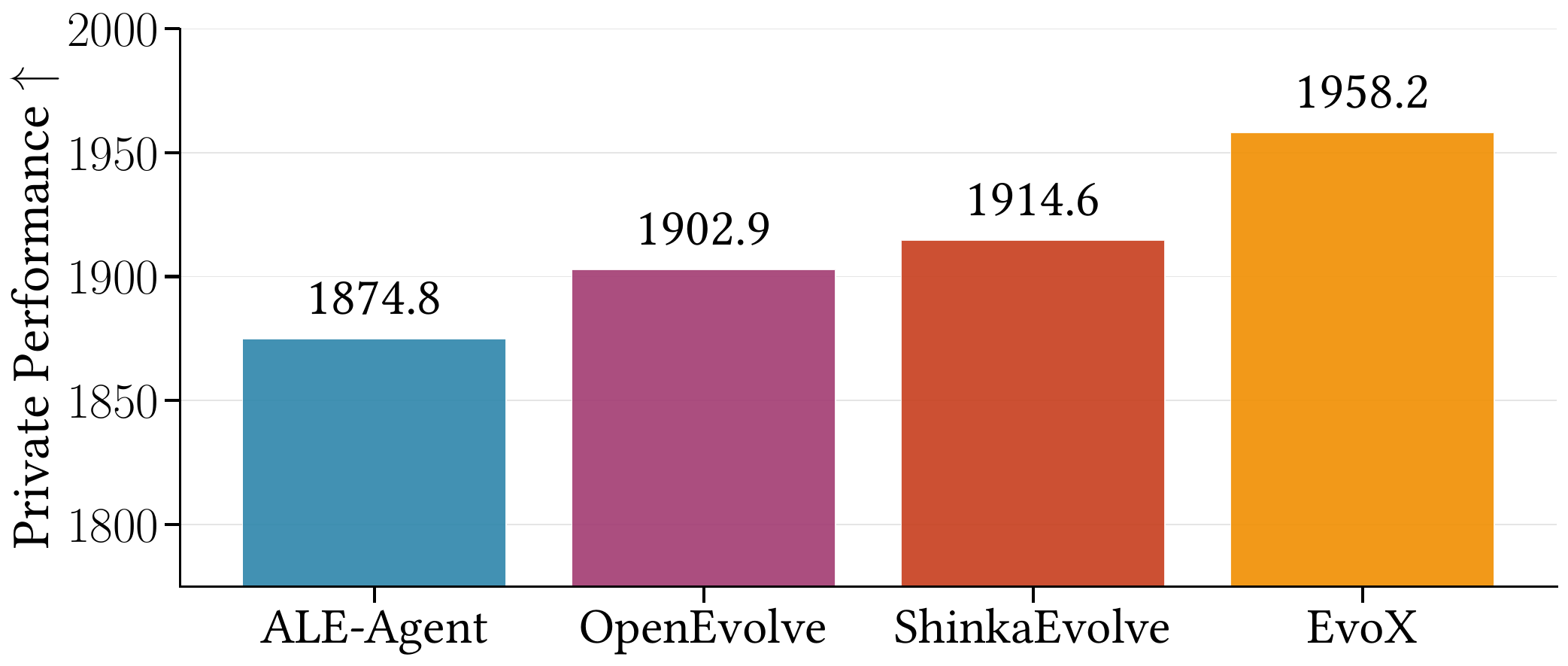}
        \caption{Average performance across 10 ALE-Bench-Lite tasks.}
        \label{fig:ale-bench-b}
    \end{subfigure}
    \hfill
    \begin{subfigure}[t]{.48\columnwidth}
        \centering
        \includegraphics[width=\linewidth]{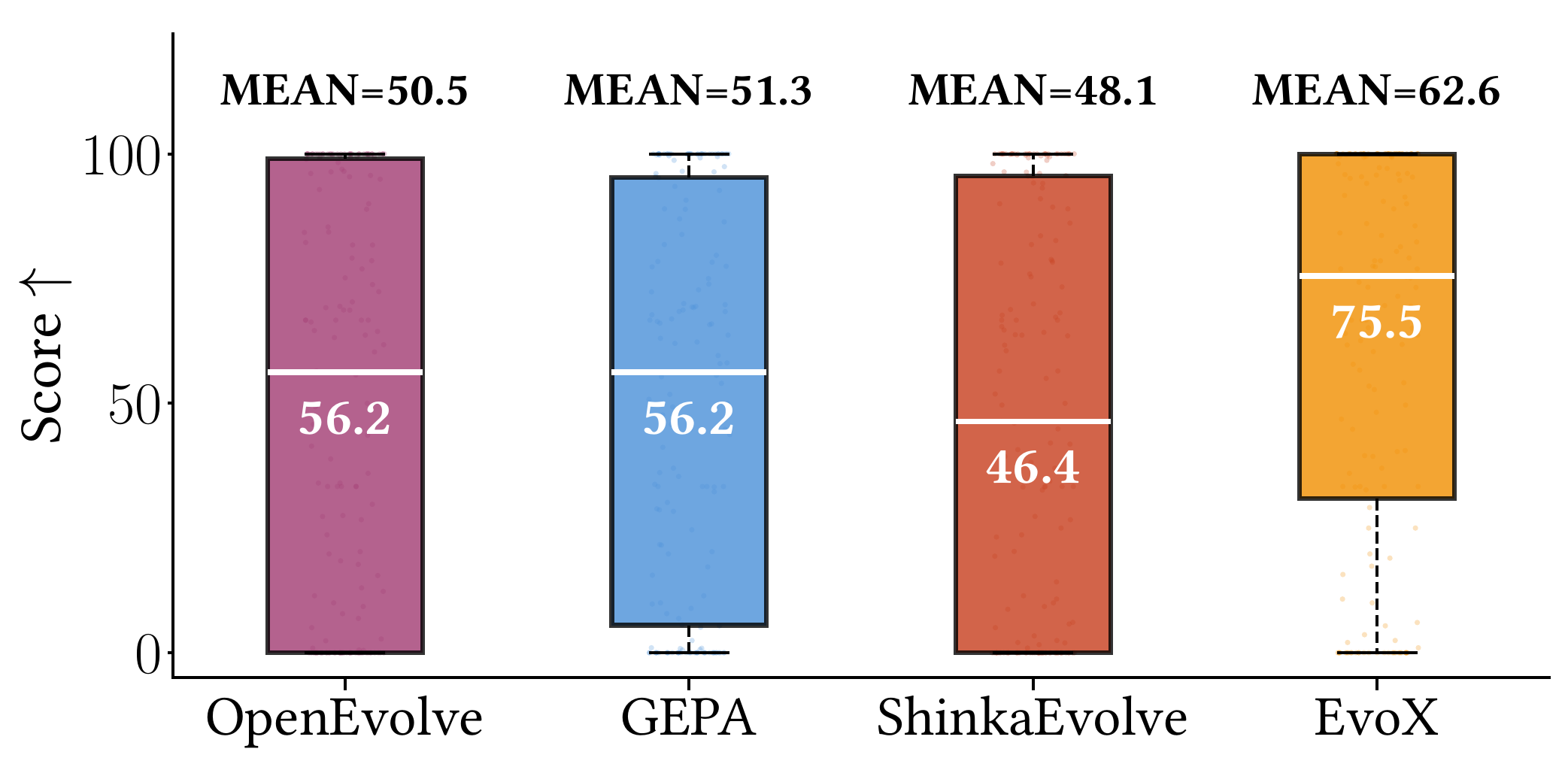}
        \caption{Performance on 172 Frontier-CS tasks.}
        \label{fig:frontierCS}
    \end{subfigure}
    \caption{
    \textbf{Algorithm and Research Challenges.}
    \sys achieves the highest average private performance with GPT-5 on 10 different ALE-Bench-Lite problems and highest median score across 172 different Frontier-CS challenges.
    }
    \label{fig:algorithm-and-research-problems}
\end{figure}

We also evaluate \sys on algorithmic and research benchmarks (Figure~\ref{fig:algorithm-and-research-problems}).

\textbf{ALE-Bench-Lite.}
As shown in Figure~\ref{fig:ale-bench-b}, we evaluate \sys on 10 tasks from ALE-Bench-Lite~\cite{imajuku2025ale}, derived from AtCoder Heuristic Contests.
Following prior work~\cite{shinka}, all methods are initialized from the ALE-Agent baseline and evaluated using private scores.

\sys achieves the strongest overall performance, attaining the highest average private score (1958.2), outperforming ALE-Agent (1874.8), OpenEvolve (1902.9), and our reproduction of ShinkaEvolve result (1914.6)~\cite{shinka}.
On AHC016 (graph classification), \sys finds a solution that replaces graph edit distance with block-pattern signatures, a noise-aware representation enabling efficient matching.
On AHC024 (map compression), it improves simulated annealing with boundary-aware tracking, prioritizing boundary updates to improve search progress.
Gains from \sys are smaller on some tasks (e.g., AHC025), where a strong initial solution might bias the search toward a local optimum.


\textbf{Frontier-CS.}
We evaluate \sys on 172 tasks from Frontier-CS~\cite{mang2025frontiercs}, a large-scale benchmark of open-ended computer science problems.
Each task is scored from 0–100, where 100 corresponds to the best-known human or optimal solution.
As shown in Figure~\ref{fig:frontierCS}, EvoX achieves the strongest overall performance, reaching a mean score of 62.6 and median of 75.5.
This improves over OpenEvolve by 24\% in mean and 34\% in median score, over GEPA by 22\% and 34\%, and over ShinkaEvolve by 30\% and 63\%, respectively.
These gains demonstrate that EvoX discovers higher-quality solutions while maintaining stronger consistency across diverse competitive programming tasks.



\subsection{Ablations}
\label{sec:ablations}

\begin{figure*}[t]
    \centering
    \begin{subfigure}[t]{.48\textwidth}
        \centering
        \includegraphics[width=\linewidth]{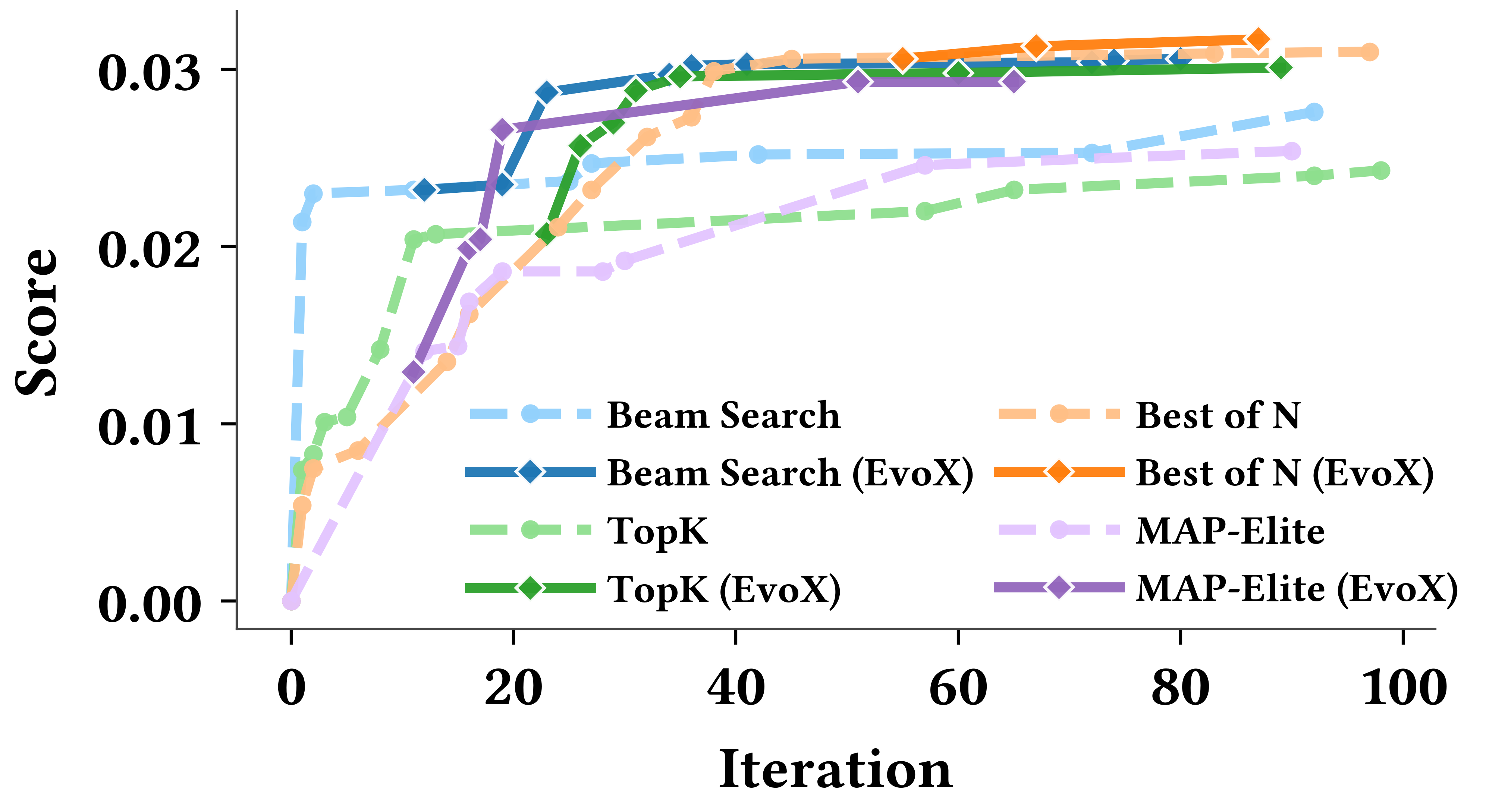}
        \caption{\textbf{Effect of different initial search strategies}}
        \label{fig:heilbronn_init}
    \end{subfigure}
    \hfill
    \begin{subfigure}[t]{.48\textwidth}
        \centering
        \includegraphics[width=\linewidth]{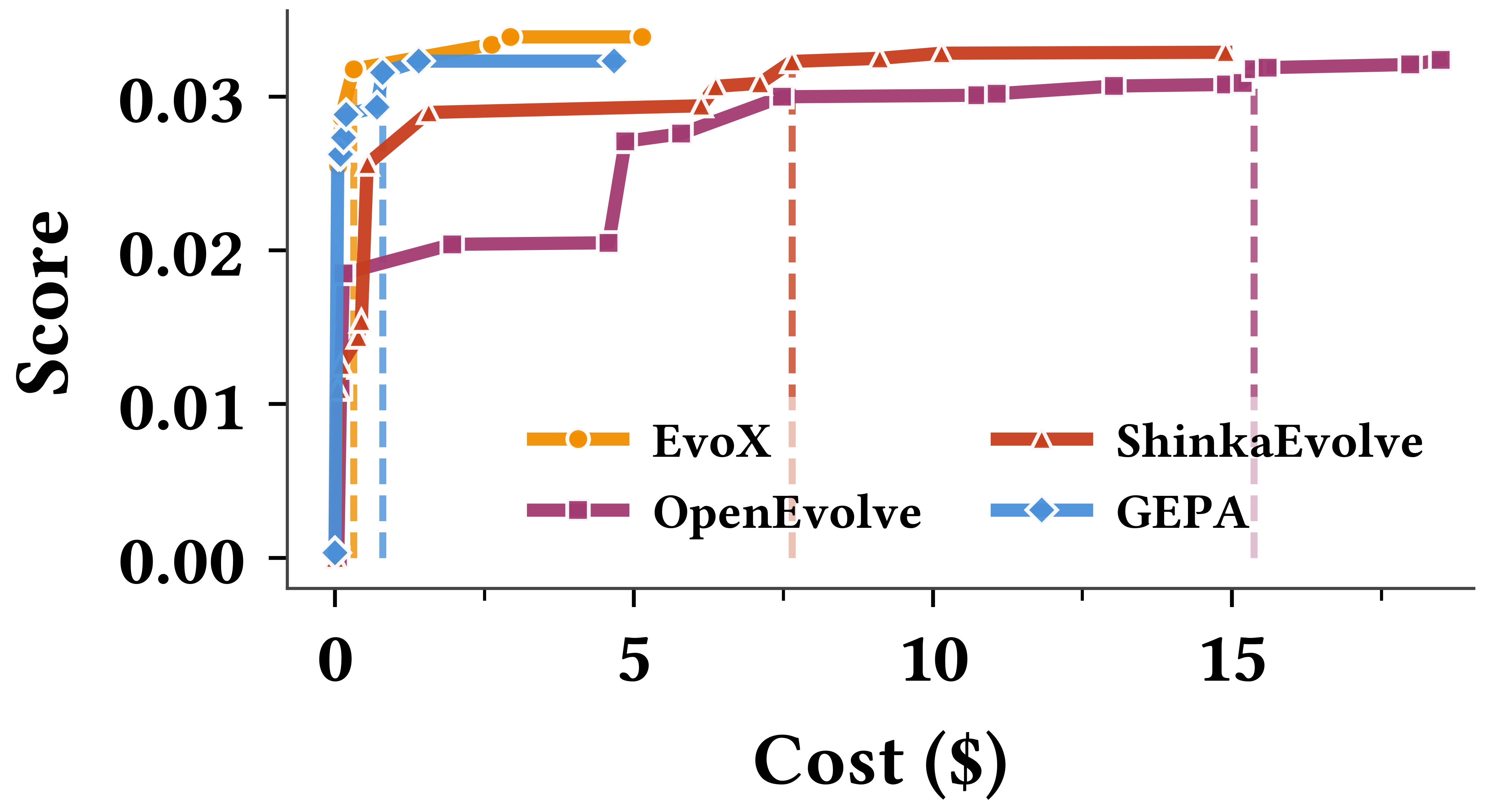}
        \caption{\textbf{Cost-quality tradeoff.}}
        \label{fig:heilbronn_cost}
    \end{subfigure}

    \caption{
    \textbf{Search strategy evolution and cost-quality tradeoffs on the Heilbronn triangle task.}
    \textbf{(a)} Dashed lines show fixed strategies, while solid lines show \sys initialized from each strategy and allowed to evolve. Regardless of initialization, \sys continues improving beyond the fixed strategy.
    \textbf{(b)} Cost-quality tradeoff under the GPT-5 model.
    To exceed a score of 0.031, \sys and GEPA both require less than \$1 in LLM generation cost, compared to ShinkaEvolve (\$7.6) and OpenEvolve (\$15.4).
    }
    \label{fig:heilbronn_combined}
\end{figure*}

\textbf{Starting from Different Search Strategies.}
Figure~\ref{fig:heilbronn_init} evaluates \sys initialized from different fixed search strategies (Beam Search, Best-of-$N$, Top-$K$, MAP-Elites) on the Heilbronn triangle task. 
The Heilbronn triangle task presents a highly non-convex objective landscape, where optimization behavior is sensitive to both global configuration discovery and subsequent local improvements.

Beam search and Best-of-$N$ provide the strongest initializations by concentrating compute on repeatedly refining top candidates: beam search via elite expansion and pruning, and Best-of-$N$ through large-batch selection. Top-$K$ instead preserves multiple candidates, diffusing optimization pressure and slowing convergence, while MAP-Elites maintains a grid of diverse solutions and lags behind.

Regardless of initialization, fixed strategies exhibit early saturation. 
In contrast, \sys consistently improves solution quality by adapting the search strategy during optimization. 
This behavior indicates robustness to initialization and demonstrates the benefits of strategy evolution.

\textbf{Cost and Scaling.}
Figure~\ref{fig:heilbronn_cost} shows the cost–quality tradeoff on the Heilbronn triangle task under the GPT-5 model. To exceed a score of 0.031, \sys and GEPA both require less than \$1 in LLM generation cost, compared to \$7.6 for ShinkaEvolve and \$15.4 for OpenEvolve. However, GEPA plateaus at a score of 0.0323 after around 20 iterations and shows no further improvement despite continued search. In contrast, \sys breaks through this stagnation point and reaches a peak score of 0.0339, which is the highest among all methods, by adapting its search strategy during optimization. This shows that \sys not only reaches competitive solutions faster, but continues to improve where fixed-strategy baselines stall.

\paragraph{Search Evolution Examples.}
We provide additional case studies of search strategy evolution in Appendix~\ref{appnd:analysis-search}.
In some tasks, strong generator models already produce high-quality candidates, allowing even simple strategies to make progress.
For example, under Gemini-3.0-Pro, random sampling alone yields substantial improvements on Cloudcast (Appendix Table~\ref{tab:appendix-adrs-random}), without specialized selection or variation.

In contrast, other tasks rely heavily on adaptive strategy.
For Signal Processing, major improvements arise during multi-objective sampling, followed by refinement-focused phases.
For Circle Packing, early free-form variation yields large gains but quickly saturates; continued progress emerges through structural variation, with later refinement contributing incremental improvements.
The Heilbronn Triangle task derives its improvements predominantly from strategies emphasizing local refinement, on top of a reasonable initial configuration identified in the early iterations.

Appendix Tables~\ref{tab:sp-search-evolution}, \ref{tab:a8}, and \ref{tab:a9} provide detailed breakdowns of these patterns.
Across tasks, both selection mechanisms and variation operators evolve in task- and run-dependent ways, reflecting differences in optimization dynamics.

%% file: table/math.tex
\begin{table*}[t]
\centering
\scriptsize
\caption{
\textbf{Main results for math optimization problems.}
We report mean and best over three runs.
“$\uparrow$” / “$\downarrow$” indicate maximization / minimization.
All open methods use 100 iterations under the same backbone model.
\textbf{Bold} denotes the best LLM-based open framework result per model.
Green cells indicate matches or improvements over human SOTA or closed-source AlphaEvolve results. In Circle Packing Rect, \sys achieves
2.36583237, exceeding AlphaEvolve’s 2.36583213, although both appear equal when rounded.
}

\vspace{1em}
\label{tab:main_results}

\setlength{\tabcolsep}{2.0pt}
\renewcommand{\arraystretch}{1.2}
\resizebox{\textwidth}{!}{
\begin{tabular}{l c c c cccc cccc}
\toprule
& & & & 
\multicolumn{4}{c}{\textbf{GPT-5}} &
\multicolumn{4}{c}{\textbf{Gemini-3.0-Pro}} \\
\cmidrule(l{8pt}r{8pt}){5-8}
\cmidrule(l{8pt}r{8pt}){9-12}

\textbf{Task} & \textbf{Stat} & \textbf{Human}
&\textbf{AlphaEvolve}
& \textbf{OpenEvolve} & \textbf{GEPA} & \textbf{Shinka} & \textbf{\sys}
& \textbf{OpenEvolve} & \textbf{GEPA} & \textbf{Shinka} & \textbf{\sys} \\
\midrule

Circle Packing ($\uparrow$)
& Mean
& -- & --
& 2.5308 & 2.6129 & 2.4642 & \textbf{2.6324}
& 2.5414 & 2.6198 & 2.6216 & \textbf{2.6328} \\
& Best
& 2.6340 & 2.6350
& 2.5414 & \cellcolor{lightgreen}\textbf{2.6359} & 2.5161 & \cellcolor{lightgreen}\textbf{2.6359}
& 2.5414 & 2.6208 & 2.6358 & \cellcolor{lightgreen}\textbf{2.6359} \\
\midrule

Circle Packing Rect ($\uparrow$)
& Mean
& -- & --
& 2.2673 & 2.3265 & 2.3347 & \textbf{2.3525}
& 2.3651 & 2.2160 & \textbf{2.3658} & \textbf{2.3658} \\
& Best
& 2.3640 & {2.3658}
& 2.2756 & 2.3537 & 2.3577 & \textbf{2.3600}
& 2.3658 & 2.2507 & \textbf{2.3658} & \cellcolor{lightgreen}\textbf{2.3658} \\
\midrule

heilbronn\_convex ($\uparrow$)
& Mean
& -- & --
& 0.0230 & 0.0259 & 0.0228 & \textbf{0.0263}
& 0.0279 & 0.0228 & 0.0279 & \textbf{0.0287} \\
& Best
& 0.0306  & \cellcolor{lightgreen}{0.0309}
& 0.0267 & 0.0269 & 0.0256 & \textbf{0.0272}
& 0.0280 & 0.0274 & 0.0287 & \textbf{0.0296} \\
\midrule

heilbronn\_triangle ($\uparrow$)
& Mean
& -- & --
& 0.0250 & 0.0317 & \textbf{0.0319} & 0.0316
& 0.0331 & 0.0311 & 0.0351 & \textbf{0.0359} \\
& Best
& 0.0360 & \cellcolor{lightgreen}{0.0365}
& 0.0283 & 0.0332 & 0.0338 & \textbf{0.0339}
& 0.0350 & 0.0329 & 0.0356 & \cellcolor{lightgreen}\textbf{0.0365} \\
\midrule

min\_max\_min\_dist ($n{=}16, d{=}2$) ($\downarrow$)
& Mean
& -- & --
& 13.00 & 12.98 & 12.99 & \textbf{12.95}
& 12.96 & 12.89 & 12.89 & \textbf{12.89} \\
& Best
& 12.89 & \cellcolor{lightgreen}{12.89}
& 13.00 & {12.95} & 12.98 & \cellcolor{lightgreen}\textbf{12.89}
& \textbf{12.89} & \textbf{12.89} & \textbf{12.89} & \cellcolor{lightgreen}\textbf{12.89} \\
\midrule

min\_max\_min\_dist ($n{=}14, d{=}3$) ($\downarrow$)
& Mean
& -- & --
& 4.51 & 4.30 & \textbf{4.19} & 4.21
& 4.17 & 4.71 & \textbf{4.16} & 4.17 \\
& Best
& 4.17 & 4.17
& 4.46 & 4.18 & \textbf{4.17} & 4.18
& \textbf{4.16} & 4.59 &\textbf{4.16} & \cellcolor{lightgreen}\textbf{4.16} \\
\midrule



third\_autocorr\_ineq ($\downarrow$)
& Mean
& -- & --
& 1.4671 & 1.4768 & 1.4785 & \textbf{1.4714}
& 1.4609 & 1.4678 & 1.4595 & \textbf{1.4589} \\
& Best
& 1.4581 & \cellcolor{lightgreen}{1.4557}
& 1.4610 & 1.4758 & 1.4614 & \textbf{1.4609}
& 1.4600 & 1.4598 & 1.4578 & \textbf{1.4558} \\
\midrule

signal processing ($\uparrow$)
& Mean
& -- & --
& 0.5686 & {0.6891} & 0.4855 & \textbf{0.7106}
& 0.5525 & 0.6169 & 0.4799 & \textbf{0.7351} \\
& Best
& -- & --
& 0.6219 & {0.7057} & 0.5328 & \textbf{0.7214}
& 0.5649 & 0.6798 & 0.5049 & \cellcolor{lightgreen}{\textbf{0.7429}}\\
\bottomrule
\end{tabular}
}
\end{table*}

%% file: table/adrs.tex
\begin{table*}[t]
\centering
\scriptsize

\caption{
\textbf{Main results for system problems.}
We report mean and best scores over three runs. “$\uparrow$” denotes maximization and “$\downarrow$” minimization. All methods (except human / prior AI) are run for 100 iterations using GPT-5 and Gemini-3.0-Pro.
\textbf{Bold} denotes the best LLM-based open framework result per model.
Green cells indicate matches or improvements over human SOTA results.
}
\vspace{1em}
\label{tab:adrs_main_results}

\setlength{\tabcolsep}{5pt}
\renewcommand{\arraystretch}{1.25}

\begin{tabular}{l c c cccc cccc}
\toprule
& & &
\multicolumn{4}{c}{\textbf{GPT-5}} &
\multicolumn{4}{c}{\textbf{Gemini-3.0-Pro}} \\
\cmidrule(l{8pt}r{8pt}){4-7}
\cmidrule(l{8pt}r{8pt}){8-11}

\textbf{Task} & \textbf{Stat} & \textbf{Human}
& \textbf{OpenEvolve} & \textbf{GEPA} & \textbf{Shinka} & \textbf{\sys}
& \textbf{OpenEvolve} & \textbf{GEPA} & \textbf{Shinka} & \textbf{\sys} \\
\midrule

EPLB ($\uparrow$)
& Mean
& --
& 0.1300 & 0.1338 & 0.1185 & \textbf{0.1358}
& 0.1272 & 0.1268 & 0.1206 & \textbf{0.1392} \\
& Best
& 0.1265
& 0.1272 & 0.1445 & 0.1272 & \cellcolor{lightgreen}\textbf{0.1453}
& 0.1272 & 0.1272 & 0.1272 & \cellcolor{lightgreen}\textbf{0.1453} \\
\midrule

PRISM ($\uparrow$)
& Mean
& --
& 25.15 & 26.19 & 26.26 & \textbf{27.67}
& 26.24 & 26.16 & 26.25 & \textbf{26.26} \\
& Best
& 21.89
& 26.23 & 26.23 & 26.26 & \cellcolor{lightgreen}\textbf{30.52}
& 26.24 & 26.19 & 26.26 & \cellcolor{lightgreen}\textbf{26.26} \\
\midrule

LLM-SQL ($\uparrow$)
& Mean
& --
& 0.7053 & 0.7127 & 0.7123 & \textbf{0.7231}
& 0.7251 & 0.7129 & 0.7210 & \textbf{0.7278} \\
& Best
& 0.6920
& 0.7155 & 0.7129 & 0.7125 & \cellcolor{lightgreen}\textbf{0.7298}
& 0.7258 & 0.7134 & 0.7212 & \cellcolor{lightgreen}\textbf{0.7300} \\
\midrule

Cloudcast ($\downarrow$)
& Mean
& --
& 851.72 & 689.89 & 954.79 & \textbf{662.26}
& 707.82 & 720.38 & 949.79 & \textbf{623.69} \\
& Best
& 626.24
& 729.80 & 645.72 & 812.74 & \textbf{637.14}
& 707.82 & 667.06 & 1032.42 & \cellcolor{lightgreen}\textbf{623.69} \\
\midrule

Transaction ($\uparrow$)
& Mean
& --
& 3860.1 & 3752.6 & 4090 & \textbf{4292.32}
& 4109.19 & 3615.57 & 3931.67 & \textbf{4267.75} \\
& Best
& 2724.8
& 4237.3 & 3984.1 & 4329 & \cellcolor{lightgreen}\textbf{4347.83}
& 4273.50 & 3615.57 & 4255.32 & \cellcolor{lightgreen}\textbf{4310.34} \\
\midrule

Telemetry Repair ($\uparrow$)
& Mean
& --
& 0.9031 & 0.9158 & 0.9229 & \textbf{0.9464}
& \textbf{0.9541} & 0.8505 & 0.9176 & 0.9381 \\
& Best
& 0.8222
& 0.9515 & 0.9477 & 0.9515 & \cellcolor{lightgreen}\textbf{0.9520}
& \textbf{0.9541} & 0.8553 & 0.9331 & \cellcolor{lightgreen}\textbf{0.9467} \\
\bottomrule
\end{tabular}
\end{table*}

%% file: tex/conclusion.tex
\section{Conclusion}
We introduced \sys, a meta-evolution method that jointly evolves candidate solutions and the search strategies that generate them. By adapting search strategy to the structure and stage of the optimization process, \sys delivers consistent improvements in solution quality and cost efficiency across diverse domains. As optimization problems continue to scale in size and complexity, these results point toward a future of general-purpose, self-evolving systems that continuously refine how they search, reducing reliance on fixed, manually designed procedures.


\section*{Acknowledgments} 
This research is supported by NSF (IFML) CCF-2019844 and gifts from Accenture, AMD,
Anyscale, Broadcom Inc., Google, IBM, Intel, Intesa Sanpaolo, Lambda, Mibura Inc, Samsung SDS,
and SAP.

%% file: tex/appendix.tex
\newpage
\begin{center}
     \Large\textbf{Appendix}
\end{center}
\label{appendix}



\input{tex/appendix/benchmark_description}
\input{tex/appendix/other_results}

\input{tex/prompt}
\newpage
\section{Analysis of Search Evolution}
\label{appnd:analysis-search}
\input{table/serach_analysis}

\input{tex/case_study}

\input{tex/appendix/best_program}
\input{tex/appendix/config}

%% file: tex/appendix/benchmark_description.tex
\section{Benchmark Details}
\label{appendix:benchmark}


We evaluate \sys on 196 problems: 24 optimization problems spanning three domains (mathematical optimization (8 problems), system performance (6 problems), and algorithmic challenges (10 problems)) and 172 algorithmic problems from FrontierCS.

\subsection{Mathematical Optimization}
\begin{table}[H]
\centering
\small
\setlength{\tabcolsep}{6pt}
\renewcommand{\arraystretch}{1.25}
\caption{\textbf{Mathematical optimization benchmarks.}
Summary of optimization tasks used to evaluate \sys, including geometric, combinatorial, and signal processing problems.}
\vspace{0.5em}
\label{tab:benchmark_math}

\begin{tabularx}{\linewidth}{p{4.8cm} p{3.0cm} X}
\toprule
\textbf{Problem} & \textbf{Objective} & \textbf{Description} \\
\midrule

Circle Packing (Square)
& $\max\, r$
& Place $n$ equal-radius circles inside the unit square without overlap; maximize the common radius subject to non-overlap and boundary constraints. \\

Circle Packing (Rectangle)
& $\max\, r$
& Pack $n$ equal-radius circles into a rectangle with fixed aspect ratio, maximizing the radius under geometric constraints. \\

Heilbronn Triangle
& $\max\, \min$ area
& Place $n$ points in $[0,1]^2$ to maximize the area of the smallest triangle formed by any three points. \\

Heilbronn Convex
& $\max\, \min$ area
& Generalization of the Heilbronn triangle problem: maximize the minimum area of convex hulls formed by any subset of $k>3$ points. \\

Min--Max--Min Distance (2 variants)
& $\max\, \frac{d_{\min}}{d_{\max}}$
& Place points to maximize uniformity, measured by the ratio between minimum and maximum pairwise distances. \\

Third Autocorrelation Inequality
& $\min\, C_3$
& Construct witness functions that minimize the third autocorrelation constant in additive combinatorics. \\

Signal Processing
& multi-objective
& Design a causal, online filtering program for noisy time series, balancing fidelity, smoothness, lag, and false trend detection. \\

\bottomrule
\end{tabularx}
\end{table}

\subsection{System Performance Optimization}

\begin{table}[H]
\centering
\small
\setlength{\tabcolsep}{6pt}
\renewcommand{\arraystretch}{1.25}
\caption{\textbf{System performance optimization benchmarks.}
Benchmarks drawn from ADRS-Bench~\cite{cheng2025barbarians}, capturing realistic optimization problems from production systems.}
\vspace{0.5em}
\label{tab:benchmark_system}

\begin{tabularx}{\linewidth}{p{4.8cm} p{3.0cm} X}
\toprule
\textbf{Problem} & \textbf{Objective} & \textbf{Description} \\
\midrule

EPLB (Expert Placement Load Balancing)
& $\max$ throughput ratio
& Place and replicate experts in MoE models across GPUs based on activation frequency, minimizing load imbalance and computational skew. \\

PRISM (Global Model Scheduling)
& $\max$ throughput
& Schedule multiple deep learning models across distributed GPUs under memory, compute, and latency SLO constraints in a multi-tenant setting. \\

LLM--SQL
& $\max$ accuracy
& Optimize LLM-driven SQL generation via prompting or post-processing; a query is correct only if it exactly matches ground-truth execution results. \\

Cloudcast (Multi-Region Data Transfer)
& $\min$ cost
& Design cost-efficient data transfer strategies across cloud regions with heterogeneous bandwidths, latencies, and egress pricing under deadline constraints. \\

Transaction Scheduling
& $\max$ commits/sec
& Schedule database transactions under contention to maximize throughput while minimizing aborts from conflicts, deadlocks, or timeouts. \\

Telemetry Repair
& $\max$ reconstruction accuracy
& Reconstruct missing or corrupted telemetry signals using partial observations, exploiting temporal and spatial correlations. \\

\bottomrule
\end{tabularx}
\end{table}

\subsection{Algorithmic and Research Problems}

\begin{table}[H]
\centering
\small
\setlength{\tabcolsep}{6pt}
\renewcommand{\arraystretch}{1.25}
\caption{\textbf{Algorithmic and research problem benchmarks.}
We evaluate on 182 open-ended problems: 10 NP-hard optimization problems from ALE-Bench-Lite~\cite{imajuku2025ale} and 172 problems from FrontierCS~\cite{mang2025frontiercs}.}
\vspace{0.5em}
\label{tab:benchmark_algorithmic_research}

\begin{tabularx}{\linewidth}{p{4.2cm} p{1.0cm} X}
\toprule
\textbf{Problem / Category} & \textbf{Count} & \textbf{Description} \\
\midrule

\multicolumn{3}{l}{\textit{ALE-Bench-Lite: NP-hard optimization from AtCoder Heuristic Contests}} \\
\addlinespace[2pt]

\texttt{ahc008} (Territory)
& 1
& Control agents on a grid to place fences isolating pets; $\max$ satisfaction. \\

\texttt{ahc011} (Sliding Puzzle)
& 1
& Rearrange tiles so line patterns form a large connected tree; $\max$ connectivity. \\

\texttt{ahc015} (Candy Clustering)
& 1
& Cluster candies of the same flavor using global tilt operations; $\max$ clustering score. \\

\texttt{ahc016} (Graph Classification)
& 1
& Design reference graphs and classify noisy, permuted query graphs; $\min$ classification error. \\

\texttt{ahc024} (Map Compression)
& 1
& Compress a grid map preserving district adjacency; $\min$ map size. \\

\texttt{ahc025} (Weight Balancing)
& 1
& Partition items with unknown weights into balanced groups; $\min$ imbalance. \\

\texttt{ahc026} (Box Stacking)
& 1
& Rearrange numbered boxes across stacks to a target configuration; $\min$ moves. \\

\texttt{ahc027} (Cleaning Route)
& 1
& Design a cyclic robot route to minimize steady-state dirt; $\min$ average dirtiness. \\

\texttt{ahc039} (Fishing Net)
& 1
& Construct a rectilinear polygon enclosing targets under a perimeter budget; $\max$ net score. \\

\texttt{ahc046} (Skating with Blocks)
& 1
& Visit target squares in order using moves, slides, and blocks; $\min$ actions. \\

\midrule

\multicolumn{3}{l}{\textit{FrontierCS: open-ended CS problems}} \\
\addlinespace[2pt]

\texttt{Optimization}
& 38
& Maximize or minimize a quantitative objective over a parameterized search space under resource constraints. IDs: 1, 2, 15, 16, 22, 25, 26, 27, 28, 30, 33, 35, 36, 40, 41, 42, 43, 44, 45, 46, 47, 48, 50, 52, 53, 54, 57, 58, 59, 61, 64, 68, 69, 79, 81, 86, 93, 229. \\

\texttt{Constructive}
& 62
& Synthesize a valid structured object (e.g., packing, graph, expression) under global constraints. IDs: 0, 3, 4, 5, 6, 7, 8, 9, 13, 17, 23, 24, 60, 62, 63, 70, 72, 73, 75, 77, 80, 82, 83, 85, 87, 89, 142, 174--193, 192, 193, 203, 205, 207, 209--214, 217, 220, 222, 225, 227, 228, 239, 241. \\

\texttt{Interactive}
& 72
& Solve a hidden-instance task via an adaptive query-response protocol, minimizing queries or interaction steps. IDs: 10, 11, 14, 101, 104, 106--113, 117, 119--125, 127, 132--135, 137, 138, 140, 141, 143--145, 147--171, 226, 231, 233, 243, 245, 247--249, 252--258. \\

\bottomrule
\end{tabularx}
\end{table}

%% file: tex/appendix/other_results.tex
\newpage

\section{Additional Results}
\label{appendix:full-results}

We report additional evaluations to examine EvoX behavior beyond the primary optimization benchmarks discussed in the main paper.

\subsection{ARC-AGI Evaluation}
\label{appendix:additional-benchmarks}

\textbf{ARC-AGI-2 Tasks}~\cite{chollet2025arc} evaluate abstract and compositional reasoning across programmatic problem-solving instances. 
Although ARC-AGI-2 is not explicitly designed as an optimization benchmark, it provides a useful testbed for analyzing cross-domain robustness.

Experiments follow the evaluation protocol described in Section~4.1. 
OpenEvolve (OE) and EvoX operate under a matched inference budget (30 LLM iterations) per task.

\begin{table*}[h]
\small
\centering
\caption{EvoX performance on ARC-AGI benchmarks. Values denote final accuracy.}
\label{tab:additional-benchmarks}

\begin{tabular}{l cc}
\toprule
Backbone & OpenEvolve & EvoX \\
\midrule
GPT-5 & 41\% & 48\% \\
Gemini-3-Pro & 45\% & 51\% \\
\bottomrule
\end{tabular}
\end{table*}

\vspace{0.5em}
\noindent
These results indicate that EvoX maintains performance improvements even on reasoning-oriented tasks outside traditional optimization settings. 
We emphasize that ARC-AGI fundamentally differs from the evolutionary optimization regime, as standard ARC evaluation assumes strict train–test separation, whereas evolutionary frameworks typically adapt during inference. 

\newpage
\subsection{Full Results}
\label{appendix:full-random-otherbaseline-results}

We additionally compare against CodeEvolve~\cite{assumpccao2025codeevolve} and ThetaEvolve~\cite{theta}, two recent systems representing strong prior approaches for LLM-driven optimization.

Across tasks, EvoX maintains consistent performance advantages over both fixed-policy and adaptive-policy baselines. Detailed per-task breakdowns are provided below.

\input{table/math_full_results}
\input{table/adrs_full_results}

%% file: table/math_full_results.tex
\begin{table*}[h]
\centering
\scriptsize
\caption{
{Main results for math optimization problems.}
We report mean and best over three runs.
“$\uparrow$” / “$\downarrow$” indicate maximization / minimization.
}
\label{appendix:table-random}

\vspace{0.6em}
\label{tab:main_results_random}

\setlength{\tabcolsep}{2pt}
\renewcommand{\arraystretch}{1.12}

\resizebox{\textwidth}{!}{%
\begin{tabular}{l c c c c c ccccc ccccc}
\toprule
& & & & & &
\multicolumn{5}{c}{{GPT-5}} &
\multicolumn{5}{c}{{Gemini-3.0-Pro}} \\
\cmidrule(l{6pt}r{6pt}){7-11}
\cmidrule(l{6pt}r{6pt}){12-16}

{Task} & {Stat} & {Human}
& {AlphaEvolve} & {ThetaEvolve} & {CodeEvolve}
& {Random} & {OpenEvolve} & {GEPA} & {Shinka} & {\sys}
& {Random} & {OpenEvolve} & {GEPA} & {Shinka} & {\sys} \\
\midrule

Circle Packing ($\uparrow$)
& Mean
& -- & -- & -- & --
& 2.4905 & 2.5308 & 2.6129 & 2.4642 & {2.6324}
& 2.5564 & 2.5414 & 2.6198 & 2.6216 & {2.6328} \\
& Best
& 2.6340 & 2.6350 & 2.6360 & 2.6360
& 2.5000 & 2.5414 & 2.6285 & 2.5078 & {2.6360}
& 2.5853 & 2.5414 & 2.6208 & 2.6358 & {2.6360} \\
\midrule

Circle Packing Rect ($\uparrow$)
& Mean
& -- & -- & -- & --
& 2.2105 & 2.2673 & 2.3265 & 2.3347 & {2.3525}
& 2.3432 & 2.3651 & 2.2160 & {2.3658} & {2.3658} \\
& Best
& 2.3640 & {2.3658} & -- & 2.3658
& 2.2783 & 2.2756 & 2.3537 & 2.3577 & {2.3600}
& 2.3652 & 2.3658 & 2.2507 & {2.3658} & {2.3658} \\
\midrule

heilbronn\_convex ($\uparrow$)
& Mean
& -- & -- & -- & --
& 0.0185 & 0.0230 & 0.0259 & 0.0228 & {0.0263}
& 0.0285 & 0.0279 & 0.0228 & 0.0279 & {0.0287} \\
& Best
& 0.0306  & {0.0309} & -- & --
& 0.0198 & 0.0267 & 0.0269 & 0.0256 & {0.0272}
& 0.0298 & 0.0280 & 0.0274 & 0.0287 & {0.0296} \\
\midrule

heilbronn\_triangle ($\uparrow$)
& Mean
& -- & -- & -- & --
& 0.0261 & 0.0250 & 0.0317 & {0.0319} & 0.0316
& 0.0318 & 0.0331 & 0.0311 & 0.0351 & {0.0359} \\
& Best
& 0.0360 & {0.0365} & -- & --
& 0.0285 & 0.0283 & 0.0332 & 0.0338 & {0.0339}
& 0.0329 & 0.0350 & 0.0329 & 0.0356 & {0.0365} \\
\midrule

min\_max\_min\_dist ($n{=}16, d{=}2$) ($\downarrow$)
& Mean
& -- & -- & -- & --
& 13.00 & 13.00 & 12.98 & 12.99 & {12.95}
& 12.92 & 12.96 & 12.89 & 12.89 & {12.89} \\
& Best
& 12.89 & {12.89} & -- & 12.89
& 13.00 & 13.00 & {12.95} & 12.98 & {12.89}
& 12.89 & {12.89} & {12.89} & {12.89} & {12.89} \\
\midrule

min\_max\_min\_dist ($n{=}14, d{=}3$) ($\downarrow$)
& Mean
& -- & -- & -- & --
& 4.47 & 4.51 & 4.19 & {4.19} & 4.21
& 4.19 & 4.17 & 4.71 & {4.16} & 4.17 \\
& Best
& 4.17 & 4.17 & -- & 4.17
& 4.46 & 4.46 & 4.17 & {4.17} & 4.18
& 4.16 & {4.16} & 4.59 & {4.16} & {4.16} \\
\midrule

third\_autocorr\_ineq ($\downarrow$)
& Mean
& -- & -- & -- & --
& 1.4820 & 1.4671 & 1.4768 & 1.4785 & {1.4714}
& 1.4594 & 1.4609 & 1.4678 & 1.4595 & {1.4589} \\
& Best 
& 1.4581 & {1.4557} & 1.4930 & --
& 1.4607 & 1.4610 & 1.4758 & 1.4614 & {1.4609}
& 1.4561 & 1.4600 & 1.4598 & 1.4578 & {1.4558} \\
\midrule

signal processing ($\uparrow$)
& Mean
& -- & -- & -- & --
& 0.5484 & 0.5686 & 0.6891 & 0.4855 & {0.6580}
& 0.5527 & 0.5525 & 0.6169 & 0.4799 & {0.7351} \\
& Best
& -- & -- & -- & --
& 0.5858 & 0.6219 & 0.7057 & 0.5328 & {0.6898}
& 0.5647 & 0.5649 & 0.6798 & 0.5049 & {{0.7429}}\\
\bottomrule
\end{tabular}%
}
\end{table*}

%% file: table/adrs_full_results.tex

\begin{table*}[h]
\centering
\scriptsize

\caption{
{Main results for system problems.}
We report mean and best scores over three runs. “$\uparrow$” denotes maximization and “$\downarrow$” minimization.
}
\vspace{0.6em}
\label{tab:appendix-adrs-random}

\setlength{\tabcolsep}{3pt}
\renewcommand{\arraystretch}{1.12}

\resizebox{\textwidth}{!}{%
\begin{tabular}{l c c ccccc ccccc}
\toprule
& & &
\multicolumn{5}{c}{{GPT-5}} &
\multicolumn{5}{c}{{Gemini-3.0-Pro}} \\
\cmidrule(l{6pt}r{6pt}){4-8}
\cmidrule(l{6pt}r{6pt}){9-13}

{Task} & {Stat} & {Human}
& {Random} & {OpenEvolve} & {GEPA} & {Shinka} & {\sys}
& {Random} & {OpenEvolve} & {GEPA} & {Shinka} & {\sys} \\
\midrule

EPLB ($\uparrow$)
& Mean
& --
& 0.1338 & 0.1300 & 0.1338 & 0.1185 & {0.1358}
& 0.1280 & 0.1272 & 0.1268 & 0.1206 & {0.1392} \\
& Best
& 0.1265
& 0.1445 & 0.1272 & 0.1445 & 0.1272 & {0.1453}
& 0.1285 & 0.1272 & 0.1272 & 0.1272 & {0.1453} \\
\midrule

PRISM ($\uparrow$)
& Mean
& --
& 26.23 & 25.15 & 26.19 & 26.26 & {27.67}
& 26.23 & 26.24 & 26.16 & 26.25 & {26.26} \\
& Best
& 21.89
& 26.23 & 26.23 & 26.23 & 26.26 & {30.52}
& 26.23 & 26.24 & 26.19 & 26.26 & {26.26} \\
\midrule

LLM-SQL ($\uparrow$)
& Mean
& --
& 0.7201 & 0.7053 & 0.7127 & 0.7123 & {0.7231}
& 0.7380 & 0.7251 & 0.7129 & 0.7210 & {0.7278} \\
& Best
& 0.6920
& 0.7289 & 0.7155 & 0.7129 & 0.7125 & {0.7298}
& 0.7440 & 0.7258 & 0.7134 & 0.7212 & {0.7300} \\
\midrule

Cloudcast ($\downarrow$)
& Mean
& --
& 659.00 & 851.72 & 689.89 & 954.79 & {662.26}
& 629.84 & 707.82 & 720.38 & 949.79 & {623.69} \\
& Best
& 626.24 
& 644.84 & 729.80 & 645.72 & 812.74 & {637.14}
& 635.72 & 707.82 & 667.06 & 1032.42 & {623.69} \\
\midrule

Transaction ($\uparrow$)
& Mean
& --
& 4123.87 & 3860.10 & 3752.6 & 4090.0 & {4292.32}
& 4037.14 & 4109.19 & 3615.57 & 3931.67 & {4267.75} \\
& Best
& 2724.8
& 4149.38 & 4237.3 & 3984.1 & 4329.0 & {4347.83}
& 4237.29 & 4273.50 & 3615.57 & 4255.32 & {4310.34} \\
\midrule

Telemetry Repair ($\uparrow$)
& Mean
& --
& 0.8959 & 0.9031 & 0.9158 & 0.9229 & {0.9464}
& 0.9179 & {0.9541} & 0.8505 & 0.9176 & 0.9381 \\
& Best
& 0.8222
& 0.9498 & 0.9515 & 0.9477 & 0.9515 & {0.9520}
& 0.9515 & {0.9541} & 0.8553 & 0.9331 & {0.9467} \\
\bottomrule
\end{tabular}%
}
\end{table*}

%% file: tex/prompt.tex
\newpage
\section{Prompt}

Below is the prompt used for search strategy evolution, together with example mutation operator labels.

\begin{configblock}{System Model: LLM-Driven Optimization with Evolving Search}

You are an expert developer implementing a search strategy for LLM-driven program optimization.
Your task is to implement \texttt{EvolvedProgramDatabase}, which realizes the
\texttt{SearchStrategy} interface used by the optimizer.

\medskip
The optimizer maintains a population database of evaluated candidate programs.
At each step, the search strategy constructs the next-generation LLM input by choosing:
(i) a parent program to modify,
(ii) a variation operator that specifies how the parent should be modified, and
(iii) an optional inspiration set of past programs drawn from the population.
The generated candidate is then evaluated and added back to the population.

\medskip
The quality of a search strategy is determined by how effectively it improves the best solution
over the course of the optimization process.

\medskip
You must implement a class named \texttt{EvolvedProgramDatabase}, which defines the behavior of the
search strategy through two methods:
\begin{itemize}
  \item \texttt{add}, which controls how evaluated programs are added to the population
  \item \texttt{sample}, which selects the parent program, the variation operator applied to it,
  and an optional inspiration set of past programs.
\end{itemize}

\medskip
The base class provides \texttt{self.programs: Dict[str, EvolvedProgram]} (all stored programs),
\texttt{self.get(program\_id)} for retrieval, and predefined label constants
\texttt{self.DIVERGE\_LABEL} and \texttt{self.REFINE\_LABEL} (must not be modified).

\medskip
Each candidate program is represented as an \texttt{EvolvedProgram}.
Its fields include \texttt{id}, \texttt{code}, \texttt{metrics} (with main score
\texttt{combined\_score}), \texttt{iteration\_found}, \texttt{timestamp},
\texttt{parent\_id}, 
\texttt{parent\_info} (a \texttt{(label, id)} tuple tracking the label applied to the parent),
and \texttt{metadata}.
These fields may be read but must not be modified.

\medskip
The \texttt{sample} method returns a tuple of two dicts:
\begin{itemize}
  \item \texttt{parent\_dict: Dict[str, EvolvedProgram]} --- maps a label to exactly one parent program;
  \item \texttt{inspiration\_set\_dict: List[EvolvedProgram]} --- lists of inspirational programs providing complementary context.
\end{itemize}

\medskip
\noindent\textbf{Label rules.}
The parent label should be the empty string \texttt{""} by default.
\texttt{self.REFINE\_LABEL} (local refinement) or \texttt{self.DIVERGE\_LABEL}
(structural variation) can also be used as the parent label according to the search states.

\medskip
Each generation step can only apply exactly one variation operator:
\begin{itemize}
  \item free-form variation (the default, label \texttt{""}),
  \item local refinement (\texttt{REFINE\_LABEL}), which requests small, structure-preserving edits,
  \item structural variation (\texttt{DIVERGE\_LABEL}), which encourages larger, exploratory changes.
\end{itemize}

\medskip
\noindent
\texttt{========================}\\
\texttt{CONSTRAINTS}\\
\texttt{========================}

\begin{enumerate}
    \item Program evaluation metrics are read-only and must not be modified.
    \item All imports and dataclass definitions must be preserved.
    \item All variables used by the strategy must be initialized with safe default values to avoid
runtime errors.
    \item When accessing metrics, always validate types before arithmetic
(\texttt{isinstance(value, (int, float))}).
\end{enumerate}

\medskip
\noindent
\texttt{========================}\\
\texttt{CRITICAL INSTRUCTION}\\
\texttt{========================}

Design a search strategy wisely and adaptively to maximize improvement.
\end{configblock}

\begin{configblock}{Search evolution — end-to-end prompt ($P_{\text{search}}$)}

\textbf{System message}
\begin{quote}
You are an expert software developer tasked with iteratively improving a codebase.
Your objective is to maximize the \textbf{combined score} while exploring diverse
solutions across feature dimensions. The system maintains a population of programs;
both high performance and diversity are valuable signals.
\end{quote}

\vspace{0.5em}

\textbf{User message}

\paragraph{Downstream problem context.}
Your search algorithm evolves solutions for the following downstream problem.
Use this information to inform your search strategy.

\begin{quote}
\texttt{\{problem\_template\}}
\end{quote}

\paragraph{Search algorithm information.}

\textbf{Search window.}  
Defines the iteration range, optimization horizon, and any adaptive
replacement rules governing the lifetime of the search algorithm.

\begin{quote}
\texttt{\{search\_window\_context\}}
\end{quote}

\textbf{Solution population statistics.}  
Summarizes the current state of the solution database, including population
size, score distribution, stagnation signals, reuse patterns, and execution traces.

\begin{quote}
\texttt{\{population\_state\}}
\end{quote}


\textbf{Current search algorithm.}  
The program to be rewritten.

\begin{quote}
\texttt{\{current\_program\}}
\end{quote}

\textbf{Relevant prior attempts.}  
Previous search-algorithm variants that may provide useful design signals.

\begin{quote}
\texttt{\{relevant\_programs\}}
\end{quote}

\paragraph{Task.}
Rewrite the program to improve the search-algorithm score given the population
state and downstream objective.

Provide the complete rewritten program and briefly explain one or two
high-level principles guiding your redesign.

Keep the implementation simple, and preserve the original program's
inputs and outputs while improving its internal behavior.

\begin{lstlisting}[language={Python}]
# Your rewritten search algorithm here
\end{lstlisting}

\end{configblock}

%% file: table/serach_analysis.tex
\begin{table*}[h]
\centering
\fontsize{8.2pt}{9.6pt}\selectfont
\setlength{\tabcolsep}{4.2pt}
\setlength{\extrarowheight}{0.7pt}
\caption{
\textbf{Search strategy evolution on Signal Processing (Gemini, 100 iterations).}
Initial score 0.499 $\rightarrow$ final 0.743.
$\Delta$: improvement achieved within the phase window; $W$: window length.
Variation operator $\pi \in \{\textit{local refinement}, \textit{structural variation}, \textit{free-form variation}\}$,
where free-form variation is the default.
}
\vspace{1em}
\label{tab:sp-search-evolution}

\begin{tabular}{l p{3.75cm} p{5.75cm} l r r}
\toprule
\textbf{Task} & \textbf{Strategy} & \textbf{Search Mechanism}
& \textbf{Variation Operator}
& $\boldsymbol{\Delta}$ & $\boldsymbol{W}$ \\
\midrule

Signal Processing
& Random sampling
& Parents and inspiration programs sampled uniformly from the population
& Free-form variation
& +0.03 & 11 \\

& Greedy sampling
& Parents sampled from top-performing candidates
& Free-form variation
& +0.01 & 10 \\

& Stratified + multi-objective sampling
& Parents selected across objective-specific score rankings
& Structural variation
& +0.12 & 19 \\

& UCB-based selection
& Usage-based parent selection with penalty for overuse
& Structural variation
& +0.06 & 16 \\

& Top-tier refinement
& Parents restricted to high-score candidates
& Local refinement
& +0.03 & 14 \\

\bottomrule
\end{tabular}
\end{table*}

\begin{table*}[h]
\centering
\fontsize{8.2pt}{9.6pt}\selectfont
\setlength{\tabcolsep}{5pt}
\setlength{\extrarowheight}{0.7pt}
\caption{
\textbf{Search strategy evolution on Circle Packing (Gemini, 100 iterations).}
Initial score 0.364 $\rightarrow$ final 1.0004.
$\Delta$: improvement achieved within the phase window; $W$: window length.
Variation operator $\pi \in \{\textit{local refinement}, \textit{structural variation}, \textit{free-form variation}\}$,
where free-form variation is the default.
}
\vspace{1em}
\label{tab:cp-search-evolution}
\begin{tabular}{l l p{4.2cm} l r r}
\toprule
\textbf{Task} & \textbf{Strategy} & \textbf{Search Mechanism}
& \textbf{Operator Bias}
& $\boldsymbol{\Delta}$ & $\boldsymbol{W}$ \\
\midrule

Circle Packing
& Random sampling
& Parents and contexts sampled uniformly from the population
& Free-form variation
& +0.59 & 21 \\

& Usage-penalized sampling
& Parent selection penalizing recently sampled candidates
& Structural variation
& +0.04 & 17 \\

& Tiered sampling
& Parents selected from elite, mid, and lower score bands
& Mixed (refinement + structural)
& +0.004 & 10 \\

& Stagnation-aware sampling
& Parent selection conditioned on recent parent-child scores
& Structural variation
& +0.003 & 10 \\

& Elite refinement
& Parents restricted to high-score candidates
& Local refinement
& 0 & 20 \\

& Quantile-biased refinement
& Parents sampled from upper quantiles of the score distribution
& Local refinement
& $\sim$2e$-$5 & 10 \\

\bottomrule
\end{tabular}
\label{tab:a8}
\end{table*}

\begin{table*}[h]
\centering
\fontsize{8.2pt}{9.6pt}\selectfont
\setlength{\tabcolsep}{3pt}
\setlength{\extrarowheight}{0.7pt}
\caption{
\textbf{Search strategy evolution on Heilbronn Triangle (Gemini, 100 iterations).}
Initial score 0.0 $\rightarrow$ final $\texttt{min\_area} \approx 0.0365$ (SOTA 1.0); best at iteration 91.
$\Delta$: improvement achieved within the phase window; $W$: window length.
Variation operator $\pi \in \{\textit{local refinement}, \textit{structural variation}, \textit{free-form variation}\}$,
where free-form variation is the default.
}
\vspace{1em}
\label{tab:triangle-search-evolution}
\begin{tabular}{l l p{4.4cm} l r r}
\toprule
\textbf{Task} & \textbf{Strategy} & \textbf{Search Mechanism}
& \textbf{Operator Bias}
& $\boldsymbol{\Delta}$ & $\boldsymbol{W}$ \\
\midrule

Heilbronn Triangle
& Stratified sampling
& Parents selected from top and mid score bands
& Free-form variation
& +0.853 & 26 \\

& Exploration-biased sampling
& Parents sampled uniformly with occasional high-score selection
& Structural variation
& 0 & 10 \\

& Usage-penalized sampling
& Parent selection penalizing recently sampled candidates
& Mixed (refinement + structural)
& +0.026 & 13 \\

& Tiered refinement
& Parents selected from top and mid score bands
& Local refinement
& +0.100 & 16 \\

& Visit-weighted sampling
& Parent selection weighted by historical sampling frequency
& Free-form variation
& 0 & 10 \\

& Refinement-focused sampling
& Parents sampled from high-score candidates
& Local refinement
& +0.021 & 10 \\

\bottomrule
\end{tabular}
\label{tab:a9}
\end{table*}

%% file: tex/case_study.tex
\newpage
\subsection{Case Study: Search Evolution in Circle Packing}
\label{sec:case-study-circle}

We now illustrate \sys through a case study on the \textbf{Circle Packing} task, where the objective is to construct a program that maximizes packing efficiency under geometric constraints. The generated program must balance multiple competing factors, including achieving dense configurations, maintaining numerical stability, preventing overlap violations, and converging reliably within the evaluation budget. Candidate programs are evaluated using a normalized packing score reflecting achieved density under constraint satisfaction.

Starting from an initial score of 0.364, \sys reaches a near-optimal score of 1.0004. Table~\ref{tab:cp-search-evolution} summarizes the sequence of evolved strategies.

\textbf{Static Baseline Behavior.}
Uniform parent sampling combined with free-form variation produces rapid early gains (+0.59), primarily by discovering coarse geometric heuristics such as greedy placement rules, collision-aware perturbations, and simple local displacement schemes. However, progress quickly saturates. Free-form variation largely generates parameter-level modifications, adjusting perturbation radii, iteration counts, or threshold values without introducing qualitatively new optimization mechanisms. Because uniform sampling increasingly selects structurally similar programs, variation operates over a narrowing region of the program space.

As a result, improvements exhibit diminishing returns. Local heuristic updates frequently induce oscillatory adjustments, repeated near-collisions, and unstable refinements near dense configurations. Constraint satisfaction remains externally enforced through penalties or rejection rules rather than intrinsically modeled by the update mechanism. Without adaptive strategy evolution, the baseline cannot redirect search toward mechanisms capable of coordinated global adjustments.

\textbf{Evolving Search Strategy in \sys.}

\textit{Exploration-Dominated Phase.}
\sys initially mirrors baseline behavior, rapidly identifying stable geometric constructions (+0.59). At this stage, improvements arise primarily from discovering viable placement schemes rather than refining precise optimization dynamics.

\textit{Diversity-Inducing Strategies.}
As structural redundancy emerges within the population, \sys evolves usage-penalized and tiered sampling strategies. These mechanisms alter search dynamics by discouraging repeated selection of recently sampled parents and explicitly sampling candidates across score bands. This transition prevents over-exploitation of early high-performing but structurally limited heuristics while increasing exposure to partially successful yet structurally diverse programs. Although immediate gains are modest (+0.04, +0.004), these phases preserve population diversity, which proves critical for subsequent mechanism discovery.

\textit{Mechanism Discovery via Structural Variation.}
Under stagnation-aware sampling, structural variation produces programs that introduce a fundamentally different optimization paradigm based on constrained numerical optimization using SLSQP. This transition represents a qualitative shift in solution construction. Earlier heuristic programs perform sequential local adjustments, modifying circle positions independently through handcrafted displacement rules. Such updates are inherently myopic and frequently induce constraint violations or oscillatory corrections. In contrast, SLSQP-based programs formulate packing as a constrained optimization problem in which circle positions become jointly optimized variables and overlap constraints are explicitly encoded.

This paradigm-level shift fundamentally alters refinement dynamics. Constraint satisfaction becomes intrinsic to the update rule rather than externally enforced, coordinated gradient-based updates mitigate oscillatory behaviors, and the optimizer implicitly captures higher-order interactions among circles. Rather than relying on independent local perturbations, updates now reflect globally coordinated corrections across multiple variables. This mechanism discovery effectively breaks prior stagnation patterns and unlocks further improvements.

\textit{Refinement-Dominated Phase.}
Once SLSQP-based solutions emerge, large structural edits increasingly destabilize high-quality configurations. \sys therefore shifts operator bias toward local refinement. Refinement now operates within a significantly stronger optimization framework, primarily adjusting convergence parameters, constraint tolerances, and step-size dynamics. Quantile-biased sampling further stabilizes search by relaxing strict elite selection, preventing overfitting to brittle local optima. Incremental gains observed in this phase reflect convergence stabilization rather than structural discovery.

This example highlights a central property of \sys: strategy evolution enables optimization-mechanism discovery rather than merely improving sampling efficiency. Early improvements arise from discovering viable geometric heuristics, whereas later improvements require identifying mechanisms capable of coordinated global updates. Static strategies typically fail to induce such transitions because variation operators alone rarely trigger paradigm shifts, uniform sampling suppresses structurally novel candidates, and exploitative refinement reinforces local heuristic biases.

By evolving strategies conditioned on both historical feedback and population-state signals, \sys identifies when refinement-based search becomes ineffective and increases structural variation pressure until new optimization mechanisms emerge. Overall, Circle Packing illustrates how adaptive strategy evolution governs phase transitions in search dynamics, enabling qualitative improvements that static search policies fail to realize.

%% file: references.bib
@article{chen2022learning,
  title={Learning to optimize: A primer and a benchmark},
  author={Chen, Tianlong and Chen, Xiaohan and Chen, Wuyang and Heaton, Howard and Liu, Jialin and Wang, Zhangyang and Yin, Wotao},
  journal={Journal of Machine Learning Research},
  volume={23},
  number={189},
  pages={1--59},
  year={2022}
}

@inproceedings{metz2019understanding,
  title={Understanding and correcting pathologies in the training of learned optimizers},
  author={Metz, Luke and Maheswaranathan, Niru and Nixon, Jeremy and Freeman, Daniel and Sohl-Dickstein, Jascha},
  booktitle={International Conference on Machine Learning},
  volume={97},
  pages={4556--4565},
  year={2019},
  organization={PMLR}
}

@article{chen2023symbolic,
  title={Symbolic discovery of optimization algorithms},
  author={Chen, Xiangning and Liang, Chen and Huang, Da and Real, Esteban and Wang, Kaiyuan and Pham, Hieu and Dong, Xuanyi and Luong, Thang and Hsieh, Cho-Jui and Lu, Yifeng and others},
  journal={Advances in neural information processing systems},
  volume={36},
  pages={49205--49233},
  year={2023}
}

@misc{metarloptimizer,
      title={Optimizing Test-Time Compute via Meta Reinforcement Fine-Tuning},
      author={Yuxiao Qu and Matthew Y. R. Yang and Amrith Setlur and Lewis Tunstall and Edward Emanuel Beeching and Ruslan Salakhutdinov and Aviral Kumar},
      year={2025},
      eprint={2503.07572},
      archivePrefix={arXiv},
      primaryClass={cs.LG},
      url={https://arxiv.org/abs/2503.07572},
}

@inproceedings{andrychowicz2016learning,
  title={Learning to learn by gradient descent by gradient descent},
  author={Andrychowicz, Marcin and Denil, Misha and Gomez, Sergio and Hoffman, Matthew W. and Pfau, David and Schaul, Tom and Shillingford, Brendan and de Freitas, Nando},
  booktitle={Advances in Neural Information Processing Systems},
  volume={29},
  pages={3981--3989},
  year={2016}
}

@techreport{openai2025gpt5,
  title={{GPT-5} System Card},
  author={{OpenAI}},
  year={2025},
  month={August},
  institution={OpenAI},
  url={https://cdn.openai.com/gpt-5-system-card.pdf}
}

@techreport{google2025gemini3,
  title={{Gemini 3 Pro} Model Card},
  author={{Google DeepMind}},
  year={2025},
  month={December},
  institution={Google DeepMind},
  url={https://storage.googleapis.com/deepmind-media/Model-Cards/Gemini-3-Pro-Model-Card.pdf}
}

@article{gepa,
  title={{GEPA}: Reflective prompt evolution can outperform reinforcement learning},
  author={Agrawal, Lakshya A and Tan, Shangyin and Soylu, Dilara and Ziems, Noah and Khare, Rishi and Opsahl-Ong, Krista and Singhvi, Arnav and Shandilya, Herumb and Ryan, Michael J and Jiang, Meng and others},
  journal={arXiv preprint arXiv:2507.19457},
  year={2025}
}

@article{alphaevolve,
  title={{AlphaEvolve}: A coding agent for scientific and algorithmic discovery},
  author={Novikov, Alexander and Vu, Ng{\^a}n and Eisenberger, Marvin and Dupont, Emilien and Huang, Po-Sen and Wagner, Adam Zsolt and Shirobokov, Sergey and Kozlovskii, Borislav and Ruiz, Francisco J. R. and Mehrabian, Abbas and others},
  journal={arXiv preprint arXiv:2506.13131},
  year={2025}
}

@article{FunSearch2024,
  title   = {Mathematical discoveries from program search with large language models},
  author  = {Romera-Paredes, Bernardino and Barekatain, Mohammadamin and Novikov, Alexander and Balog, Matej and Kumar, M. Pawan and Dupont, Emilien and Ruiz, Francisco J. R. and Ellenberg, Jordan S. and Wang, Pengming and Fawzi, Omar and Kohli, Pushmeet and Fawzi, Alhussein},
  journal = {Nature},
  volume  = {625},
  number  = {7995},
  pages   = {468--475},
  year    = {2024},
  doi     = {10.1038/s41586-023-06924-6},
  url     = {https://doi.org/10.1038/s41586-023-06924-6}
}

@article{theta,
  title={{ThetaEvolve}: Test-time learning on open problems},
  author={Wang, Yiping and Su, Shao-Rong and Zeng, Zhiyuan and Xu, Eva and Ren, Liliang and Yang, Xinyu and Huang, Zeyi and He, Xuehai and Ma, Luyao and Peng, Baolin and others},
  journal={arXiv preprint arXiv:2511.23473},
  year={2025}
}

@article{cheng2025barbarians,
  title={Barbarians at the gate: How {AI} is upending systems research},
  author={Cheng, Audrey and Liu, Shu and Pan, Melissa and Li, Zhifei and Wang, Bowen and Krentsel, Alex and Xia, Tian and Cemri, Mert and Park, Jongseok and Yang, Shuo and others},
  journal={arXiv preprint arXiv:2510.06189},
  year={2025}
}

@article{imajuku2025ale,
  title={{ALE-Bench}: A Benchmark for Long-Horizon Objective-Driven Algorithm Engineering},
  author={Imajuku, Yuki and Horie, Kohki and Iwata, Yoichi and Aoki, Kensho and Takahashi, Naohiro and Akiba, Takuya},
  journal={arXiv preprint arXiv:2506.09050},
  year={2025}
}

@misc{shinka,
      title={{ShinkaEvolve}: Towards Open-Ended And Sample-Efficient Program Evolution}, 
      author={Robert Tjarko Lange and Yuki Imajuku and Edoardo Cetin},
      year={2025},
      eprint={2509.19349},
      archivePrefix={arXiv},
      primaryClass={cs.CL},
      url={https://arxiv.org/abs/2509.19349}, 
}

@misc{openevolve,
  title = {{OpenEvolve}: an open-source evolutionary coding agent},
  author = {Asankhaya Sharma},
  year = {2025},
  howpublished = {GitHub},
  url = {https://github.com/codelion/openevolve}
}

@misc{reflexion,
      title={Reflexion: Language Agents with Verbal Reinforcement Learning}, 
      author={Noah Shinn and Federico Cassano and Edward Berman and Ashwin Gopinath and Karthik Narasimhan and Shunyu Yao},
      year={2023},
      eprint={2303.11366},
      archivePrefix={arXiv},
      primaryClass={cs.AI},
      url={https://arxiv.org/abs/2303.11366}, 
}

@article{lee2025evolving,
  title={Evolving deeper llm thinking},
  author={Lee, Kuang-Huei and Fischer, Ian and Wu, Yueh-Hua and Marwood, Dave and Baluja, Shumeet and Schuurmans, Dale and Chen, Xinyun},
  journal={arXiv preprint arXiv:2501.09891},
  year={2025}
}

@article{yan2026pacevolve,
  title={PACEvolve: Enabling Long-Horizon Progress-Aware Consistent Evolution},
  author={Yan, Minghao and Peng, Bo and Coleman, Benjamin and Chen, Ziqi and Xie, Zhouhang and He, Zhankui and Sachdeva, Noveen and Ye, Isabella and Wang, Weili and Wang, Chi and others},
  journal={arXiv preprint arXiv:2601.10657},
  year={2026}
}

@article{yuksekgonul2026learning,
  title={Learning to discover at test time},
  author={Yuksekgonul, Mert and Koceja, Daniel and Li, Xinhao and Bianchi, Federico and McCaleb, Jed and Wang, Xiaolong and Kautz, Jan and Choi, Yejin and Zou, James and Guestrin, Carlos and others},
  journal={arXiv preprint arXiv:2601.16175},
  year={2026}
}

@article{cai2025flex,
  title={Flex: Continuous agent evolution via forward learning from experience},
  author={Cai, Zhicheng and Guo, Xinyuan and Pei, Yu and Feng, Jiangtao and Su, Jinsong and Chen, Jiangjie and Zhang, Ya-Qin and Ma, Wei-Ying and Wang, Mingxuan and Zhou, Hao},
  journal={arXiv preprint arXiv:2511.06449},
  year={2025}
}

@article{assumpccao2025codeevolve,
  title={Codeevolve: An open source evolutionary coding agent for algorithm discovery and optimization},
  author={Assump{\c{c}}{\~a}o, Henrique and Ferreira, Diego and Campos, Leandro and Murai, Fabricio},
  journal={arXiv preprint arXiv:2510.14150},
  year={2025}
}

@article{zhang2025agentic,
  title={Agentic context engineering: Evolving contexts for self-improving language models},
  author={Zhang, Qizheng and Hu, Changran and Upasani, Shubhangi and Ma, Boyuan and Hong, Fenglu and Kamanuru, Vamsidhar and Rainton, Jay and Wu, Chen and Ji, Mengmeng and Li, Hanchen and others},
  journal={arXiv preprint arXiv:2510.04618},
  year={2025}
}

@article{chollet2025arc,
  title={Arc-agi-2: A new challenge for frontier ai reasoning systems},
  author={Chollet, Francois and Knoop, Mike and Kamradt, Gregory and Landers, Bryan and Pinkard, Henry},
  journal={arXiv preprint arXiv:2505.11831},
  year={2025}
}

@article{georgiev2025mathematical,
  title={Mathematical exploration and discovery at scale},
  author={Georgiev, Bogdan and G{\'o}mez-Serrano, Javier and Tao, Terence and Wagner, Adam Zsolt},
  journal={arXiv preprint arXiv:2511.02864},
  year={2025}
}

@book{shenoi2005introduction,
  title={Introduction to digital signal processing and filter design},
  author={Shenoi, Belle A},
  volume={169},
  year={2005},
  publisher={John Wiley \& Sons}
}

@article{chopra2025feedback,
  title={Feedback-aware monte carlo tree search for efficient information seeking in goal-oriented conversations},
  author={Chopra, Harshita and Shah, Chirag},
  journal={arXiv preprint arXiv:2501.15056},
  year={2025}
}

@article{pourcel2025self,
  title={Self-improving language models for evolutionary program synthesis: A case study on ARC-AGI},
  author={Pourcel, Julien and Colas, C{\'e}dric and Oudeyer, Pierre-Yves},
  journal={arXiv preprint arXiv:2507.14172},
  year={2025}
}

@article{fernando2023promptbreeder,
  title={Promptbreeder: Self-referential self-improvement via prompt evolution},
  author={Fernando, Chrisantha and Banarse, Dylan and Michalewski, Henryk and Osindero, Simon and Rockt{\"a}schel, Tim},
  journal={arXiv preprint arXiv:2309.16797},
  year={2023}
}

@article{madaan2023self,
  title={Self-refine: Iterative refinement with self-feedback},
  author={Madaan, Aman and Tandon, Niket and Gupta, Prakhar and Hallinan, Skyler and Gao, Luyu and Wiegreffe, Sarah and Alon, Uri and Dziri, Nouha and Prabhumoye, Shrimai and Yang, Yiming and others},
  journal={Advances in neural information processing systems},
  volume={36},
  pages={46534--46594},
  year={2023}
}

@article{packer2023memgpt,
  title={MemGPT: towards LLMs as operating systems.},
  author={Packer, Charles and Fang, Vivian and Patil, Shishir\_G and Lin, Kevin and Wooders, Sarah and Gonzalez, Joseph\_E},
  year={2023},
  publisher={ArXiv}
}

@article{suzgun2025dynamic,
  title={Dynamic cheatsheet: Test-time learning with adaptive memory},
  author={Suzgun, Mirac and Yuksekgonul, Mert and Bianchi, Federico and Jurafsky, Dan and Zou, James},
  journal={arXiv preprint arXiv:2504.07952},
  year={2025}
}

@article{shojaee2024llm,
  title={Llm-sr: Scientific equation discovery via programming with large language models},
  author={Shojaee, Parshin and Meidani, Kazem and Gupta, Shashank and Farimani, Amir Barati and Reddy, Chandan K},
  journal={arXiv preprint arXiv:2404.18400},
  year={2024}
}

@article{hemberg2024evolving,
  title={Evolving code with a large language model},
  author={Hemberg, Erik and Moskal, Stephen and O’Reilly, Una-May},
  journal={Genetic Programming and Evolvable Machines},
  volume={25},
  number={2},
  pages={21},
  year={2024},
  publisher={Springer}
}

@article{ye2024reevo,
  title={Reevo: Large language models as hyper-heuristics with reflective evolution},
  author={Ye, Haoran and Wang, Jiarui and Cao, Zhiguang and Berto, Federico and Hua, Chuanbo and Kim, Haeyeon and Park, Jinkyoo and Song, Guojie},
  journal={Advances in neural information processing systems},
  volume={37},
  pages={43571--43608},
  year={2024}
}

@article{guo2023evoprompt,
  title={Evoprompt: Connecting llms with evolutionary algorithms yields powerful prompt optimizers},
  author={Guo, Qingyan and Wang, Rui and Guo, Junliang and Li, Bei and Song, Kaitao and Tan, Xu and Liu, Guoqing and Bian, Jiang and Yang, Yujiu},
  journal={arXiv e-prints},
  pages={arXiv--2309},
  year={2023}
}

@article{jiang2026deltaevolve,
  title={DeltaEvolve: Accelerating Scientific Discovery through Momentum-Driven Evolution},
  author={Jiang, Jiachen and Ding, Tianyu and Zhu, Zhihui},
  journal={arXiv preprint arXiv:2602.02919},
  year={2026}
}

@article{mang2025frontiercs,
  title={FrontierCS: Evolving Challenges for Evolving Intelligence},
  author={Mang, Qiuyang and Chai, Wenhao and Li, Zhifei and Mao, Huanzhi and Zhou, Shang and Du, Alexander and Li, Hanchen and Liu, Shu and Chen, Edwin and Wang, Yichuan and others},
  journal={arXiv preprint arXiv:2512.15699},
  year={2025}
}
